\documentclass[10pt,twocolumn,letterpaper]{article}

\usepackage{iccv}
\usepackage{times}
\usepackage{epsfig}
\usepackage{graphicx}
\usepackage{amsmath}
\usepackage{amssymb}
\usepackage{tabu} 
\newcommand\figref{Fig.~\ref}
\newcommand\blfootnote[1]{%
  \begingroup
  \renewcommand\thefootnote{}\footnote{#1}%
  \addtocounter{footnote}{-1}%
  \endgroup
}

\usepackage[pagebackref=true,breaklinks=true,letterpaper=true,colorlinks,bookmarks=false]{hyperref}

\iccvfinalcopy

\def\iccvPaperID{****}

\ificcvfinal\fi

\begin{document}

\title{NTIRE 2021 Depth Guided Image Relighting Challenge}
\author{Majed El Helou$^1$ \and
Ruofan Zhou$^1$ \and
Sabine S\"usstrunk$^1$ \and
Radu Timofte$^1$ \and
% IPCV-IITM
Maitreya Suin$^*$ \and
A. N. Rajagopalan$^*$ \and
% Wit-AI-lab
Yuanzhi Wang$^*$ \and
Tao Lu$^*$ \and
Yanduo Zhang$^*$\and
Yuntao Wu$^*$ \and
% AICSNTU
Hao-Hsiang Yang$^*$ \and
Wei-Ting Chen$^*$\and
Sy-Yen Kuo$^*$\and
Hao-Lun Luo$^*$\and
% DeepBlueAI
Zhiguang Zhang$^*$\and
Zhipeng Luo$^*$\and 
Jianye He$^*$\and
% MCG-NKU
Zuo-Liang Zhu$^*$\and
Zhen Li$^*$\and
Jia-Xiong Qiu$^*$\and
Zeng-Sheng Kuang$^*$\and
Cheng-Ze Lu$^*$\and
Ming-Ming Cheng$^*$\and
Xiu-Li Shao$^*$\and
% alphaRelighting
Chenghua Li$^*$\and
Bosong Ding$^*$\and
Wanli Qian$^*$\and
Fangya Li$^*$\and
% VUE
Fu Li$^*$\and
Ruifeng Deng$^*$\and
Tianwei Lin$^*$\and
Songhua Liu$^*$\and
Xin Li$^*$\and
Dongliang He$^*$\and
% iPAL-RelightNet
Amirsaeed Yazdani$^*$\and
Tiantong Guo$^*$\and
Vishal Monga $^*$\and
% NPU-CVPG
Ntumba Elie Nsampi $^*$\and
Zhongyun Hu $^*$\and
Qing Wang $^*$\and
% Couger AI
Sabari Nathan$^*$ \and
Priya Kansal$^*$ \and
% usuitakumi
Tongtong Zhao$^*$\and
Shanshan Zhao$^*$
}

\maketitle
\ificcvfinal\thispagestyle{empty}\fi

\begin{abstract}
Image relighting is attracting increasing interest due to its various applications. From a research perspective, image relighting can be exploited to conduct both image normalization for domain adaptation, and also for data augmentation. It also has multiple direct uses for photo montage and aesthetic enhancement. In this paper, we review the NTIRE 2021 depth guided image relighting challenge. 

We rely on the VIDIT dataset for each of our two challenge tracks, including depth information. The first track is on one-to-one relighting where the goal is to transform the illumination setup of an input image (color temperature and light source position) to the target illumination setup. In the second track, the any-to-any relighting challenge, the objective is to transform the illumination settings of the input image to match those of another guide image, similar to style transfer. In both tracks, participants were given depth information about the captured scenes. We had nearly 250 registered participants, leading to 18 confirmed team submissions in the final competition stage. The competitions, methods, and final results are presented in this paper. 
\end{abstract}

\blfootnote{Majed El Helou, Ruofan Zhou, Sabine S\"usstrunk \textit{(majed.elhelou, sabine.susstrunk)@epfl.ch}, and Radu Timofte \textit{radu.timofte@vision.ee.ethz.ch}, are the challenge organizers, and the other authors are challenge participants.\\ $^*$Appendix~\ref{sec:teams} lists all the teams and affiliations.}

\section{Introduction}
Due to its broad utility in research and in practice, image relighting is gaining increasingly more attention. An image relighting method would enable extended data augmentation, by providing images with various illumination settings, and similarly enables illumination domain adaptation as it can transform test images into a standard unique illumination setup. In practice, image relighting is also useful for photo montage and other aesthetic image retouching applications. The task of image relighting is however challenging. The relighting method needs to understand the geometry and illumination of the scene, be able to remove shadows, recast shadows, and occasionally inpaint totally dark areas, on top of the transformation of the illuminant. 

The objective of this image relighting challenge is to push forward the state-of-the-art in image relighting, and provide a benchmark to assess competing solutions. We rely on the novel dataset \textbf{V}irtual \textbf{I}mage \textbf{D}ataset for \textbf{I}llumination \textbf{T}ransfer (VIDIT)~\cite{elhelou2020vidit}, which we briefly discuss in the following section. We refer the reader to some previous solutions on this dataset~\cite{puthussery2020wdrn,gafton20202d,dherse2020scene,hu2020sa,dong2020ensemble,wang2020deep,dasmsr,das2021dsrn}, and to the first edition of this challenge~\cite{elhelou2020aim} for an extensive overview of related datasets and a more detailed discussion about the use of VIDIT. In contrast with the first edition, we exploit depth map information to improve the scene understanding of the methods and in turn improve the final results. Depth is important as it relates to different image degradations such as chromatic aberrations~\cite{llanos2020simultaneous,elhelou2018aam,zhao2020modified}, which can cause various depth-dependent blur effects that should be properly synthesized across spectral channels~\cite{el2017correlation}, especially if the solutions should be extended to multi-spectral data in future work. More importantly, on top of the importance of geometry understanding, having auxiliary information can generally improve the overall learning~\cite{elhelou2020blind} and interpretability of the solutions. This additional internal learning is exploited by the participating teams, as discussed in the following sections.

%%%
This challenge is one of the NTIRE 2021 associated challenges: nonhomogeneous dehazing~\cite{ancuti2021ntire}, defocus deblurring using dual-pixel~\cite{abuolaim2021ntire}, depth guided image relighting~\cite{elhelou2021ntire}, image deblurring~\cite{nah2021ntire}, multi-modal aerial view imagery classification~\cite{liu2021ntire}, learning the super-resolution space~\cite{lugmayr2021ntire}, quality enhancement of heavily compressed videos~\cite{yang2021ntire}, video super-resolution~\cite{son2021ntire}, perceptual image quality assessment~\cite{gu2021ntire}, burst super-resolution~\cite{bhat2021ntire}, high dynamic range~\cite{perez2021ntire}.
%%%

\section{Depth guided image relighting}
\subsection{Dataset}
All challenge tracks exploit the novel VIDIT~\cite{elhelou2020vidit} dataset that contains 300 training scenes and 90 scenes divided equally between validation set and test set, all being mutually exclusive. Scenes are each captured 40 times: from 8 equally-spaced azimuthal angles, and with 5 color temperatures for the illumination. Image resolution is $1024\times 1024$, and the full resolution is used in both tracks. Additionally, for this edition of the challenge, the associated depth maps are used by the participants. The full dataset can be found online\footnote{\url{https://github.com/majedelhelou/VIDIT}}, with the exception of the ground-truth test data that is kept private. For reporting purposes, in research papers outside of the challenge, authors typically provide their results on the validation set.

\subsection{Challenge tracks}
The tracks 1 and 2 are similar to the tracks 1 and 3 of our first edition~\cite{elhelou2020aim}, respectively, but with the addition of depth map information to guide the relighting, and the use of full $1024\times 1024$ resolution images. \\

\textbf{Track 1: One-to-one relighting.}
In this first track, the illumination settings of both the input and the output images are pre-determined and fixed. The objective is therefore to transform an image from its original illumination settings to a known output illumination setup. \\

\textbf{Track 2: Any-to-any relighting.}
The second track allows for more flexibility in the target illumination. More specifically, the target illumination settings are dictated by a guide image similar to style transfer applications. \\

\textbf{Evaluation protocol.} We evaluate the results using the standard PSNR and SSIM~\cite{wang2004image} metrics, and the self-reported run-times and implementation details are also provided in Tables~\ref{table:T1} and~\ref{table:T2}. For the final ranking, we define a Mean Perceptual Score (MPS) as the average of the normalized SSIM and LPIPS~\cite{zhang2018unreasonable} scores, themselves averaged across the entire test set of each submission
\begin{equation}\label{eq:MPS}
    0.5\cdot(S + (1-L)),
\end{equation}
where $S$ is the SSIM score, and $L$ is the LPIPS score. \\

\textbf{Challenge phases.} (1) Development phase: registered participants have access to the full training data (input and ground-truth images and depth maps), and to the input data of the validation set. A leader board enables the participants to get immediate automated feedback on their performance and their ranking relative to the other competing teams, by uploading their validation set output results to our server. (2) Testing: registered participants have access to the input of the test sets, and can upload their results in a similar way as during the development phase. However, the difference is that results are not shown to the participants to counter any potential overfitting. We only accept a final submission of the test set outputs when it is accompanied by reproducible open-source code, and a fact sheet containing all the details of the proposed solution.

\section{Challenge results}
The results of our tracks 1 and 2 are collected in Tables~\ref{table:T1} and ~\ref{table:T2}, respectively. Challenge solutions are described in the following section for each team.
We also show some visual results from top-performing solutions with the associated input, depth, and ground-truth information in Fig.~\ref{fig:track1} for track 1, and in Fig.~\ref{fig:track2} for track 2. Generally, all solutions outperform the results we had in the first edition when depth information was not used. This strongly supports the importance of depth maps for the image relighting tasks.

\begin{table*}[htp!]
\centering
\resizebox{\textwidth}{!}{%
\begin{tabular}{l|l||c|c|c|c||c|c|c}
Team&Author &\textbf{MPS} $\uparrow$&SSIM $\uparrow$&LPIPS $\downarrow$&PSNR $\uparrow$&Run-time&Platform&GPU\\
\hline\hline
AICSNTU-MBNet & HaoqiangYang & 0.7663 & 0.6931 & 0.1605 & 19.1469 & 2.88s& PyTorch &Tesla V100 \\
iPAL-RelightNet & auy200& 0.7620 & 0.6874 & 0.1634 & 18.8358 & 0.53s&PyTorch &Titan XP \\
NTUAICS-ADNet & aics & 0.7601 & 0.6799 & 0.1597 & 18.8639 & 2.76s &  PyTorch &Tesla V100 \\
VUE & lifu & 0.7600 & 0.6903 & 0.1702 & 19.8645 & 0.23s & PyTorch & P40\\
NTUAICS-VGG &jimmy3505090 & 0.7551 & 0.6772 & 0.1670 & 18.2766 & 2.12s & PyTorch &Tesla V100 \\
DeepBlueAI & DeepBlueAI & 0.7494 & 0.6879 & 0.1891 & 19.8784 & 0.17s & PyTorch & Tesla V100\\
usuitakumi & usuitakumi & 0.7229 & 0.6260 & 0.1801 & 16.8249 & 0.04s & PyTorch & Tesla V100\\
MCG-NKU & NK\_ZZL & 0.7147 & 0.6191 & 0.1896 & 19.0856 & 0.33s & PyTorch & RTX TITAN \\
alphaRelighting & lchia & 0.7101 & 0.6084 & 0.1882 & 15.8591 & 0.04s & PyTorch & Tesla K80 \\
Wit-AI-lab & MDSWYZ & 0.6966 & 0.6113 & 0.2181 & 17.5740 & 0.9s & PyTorch & RTX 2080Ti \\
Couger AI & Sabarinathan & 0.6475 & 0.5469 & 0.2518 & 18.2938 &0.015s & Tensorflow & GTX 1070 \\

%\textcolor{red}{EXAMPLE}&\textcolor{red}{codalab username}&0.6452&0.6310&0.3405&17.0717&0.03s&Tensorflow&1x P100\\

\hline\hline \\
\end{tabular}}
\caption{NTIRE 2021 Depth-Guided Image Relighting Challenge Track 1 (One-to-one relighting) results. The MPS, used to determine the final ranking, is computed following Eq.~\eqref{eq:MPS}.} %$^*$ Disqualified from track 1 awards due to a double submission by the same team (NTUAICS-VGG), when a teams makes double submissions we consider the \textit{worse-performing} one for the awards.}
\label{table:T1}
\end{table*}

\begin{table*}[htp!]
\centering
\resizebox{\textwidth}{!}{%
\begin{tabular}{l|l||c|c|c|c||c|c|c}
Team&Author &\textbf{MPS} $\uparrow$&SSIM $\uparrow$&LPIPS $\downarrow$&PSNR $\uparrow$&Run-time&Platform&GPU\\
\hline\hline
DeepBlueAI & DeepBlueAI & 0.7675 & 0.7087 & 0.1737 & 20.7915 & 0.17s & PyTorch &Tesla V100\\
VUE & lifu & 0.7671 & 0.6874 & 0.1532 & 19.8901 & 0.3s & PyTorch & P40\\
AICSNTU-SSS & HaoqiangYang & 0.7609 & 0.6784 & 0.1566 & 19.2212 & 2.04s& PyTorch &Tesla V100 \\
NPU-CVPG & elientumba & 0.7423 & 0.6508 & 0.1661 & 18.6039 & 0.674s & PyTorch & TITAN RTX \\
iPAL-RelightNet & auy200& 0.7341 & 0.6711 & 0.2028 & 20.1478 &0.51s &PyTorch &Titan XP \\
IPCV\_IITM & ms\_ipcv & 0.7172 & 0.6052 & 0.1708 & 18.7472 & 0.3s & PyTorch & TitanX\\
Wit-AI-lab & MDSWYZ & 0.6976 & 0.5985 & 0.2032 & 17.5222 & 6s & PyTorch & RTX 2070 \\

%\textcolor{red}{EXAMPLE}&\textcolor{red}{codalab username}&0.6452&0.6310&0.3405&17.0717&0.03s&Tensorflow&1x P100\\

\hline\hline \\
\end{tabular}}
\caption{NTIRE 2021 Depth-Guided Image Relighting Challenge Track 2 (Any-to-any relighting) results. The MPS, used to determine the final ranking, is computed following Eq.~\eqref{eq:MPS}.}
\label{table:T2}
\end{table*}

\newcommand{\viditfig}[1]{\includegraphics[width=0.15\linewidth]{#1}}
\begin{figure*}[htp!]
 \centering
    \begin{tabu}{cccccc}
    \viditfig{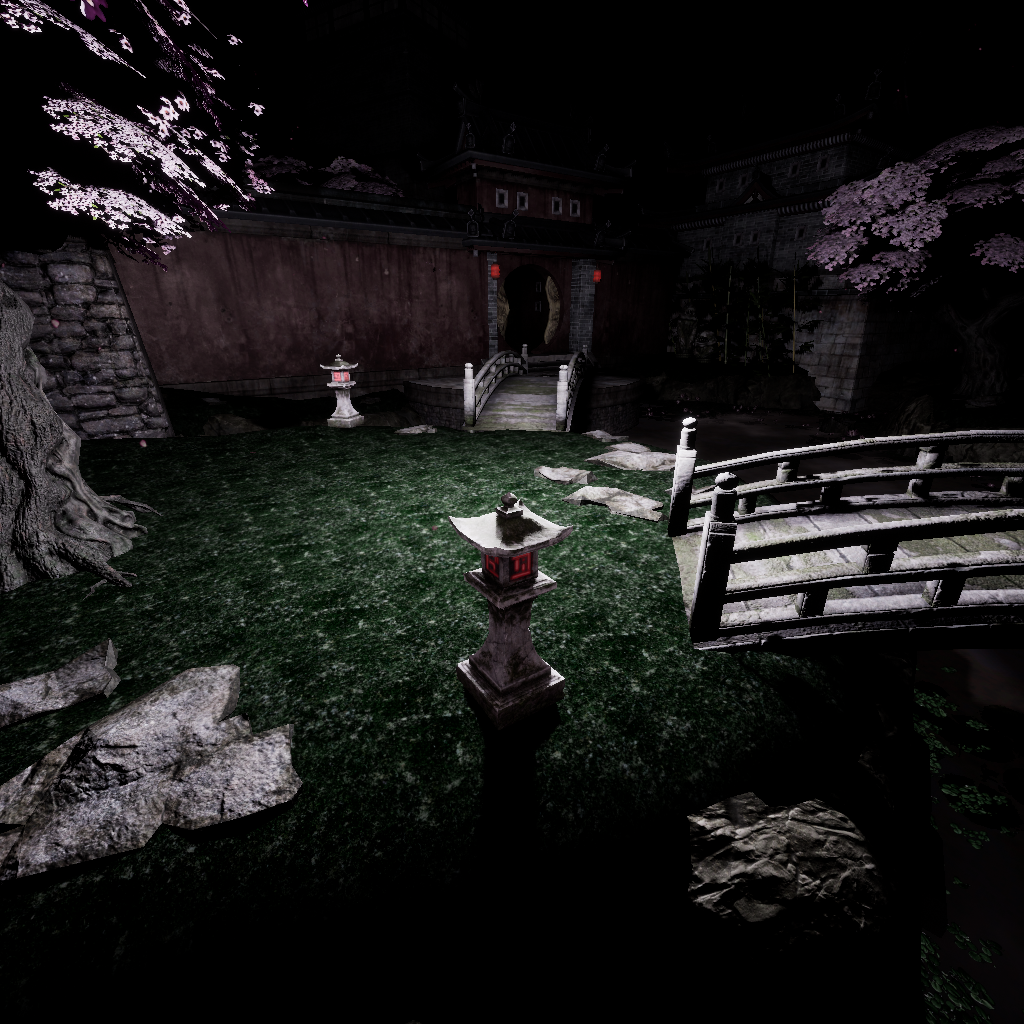} &
    \viditfig{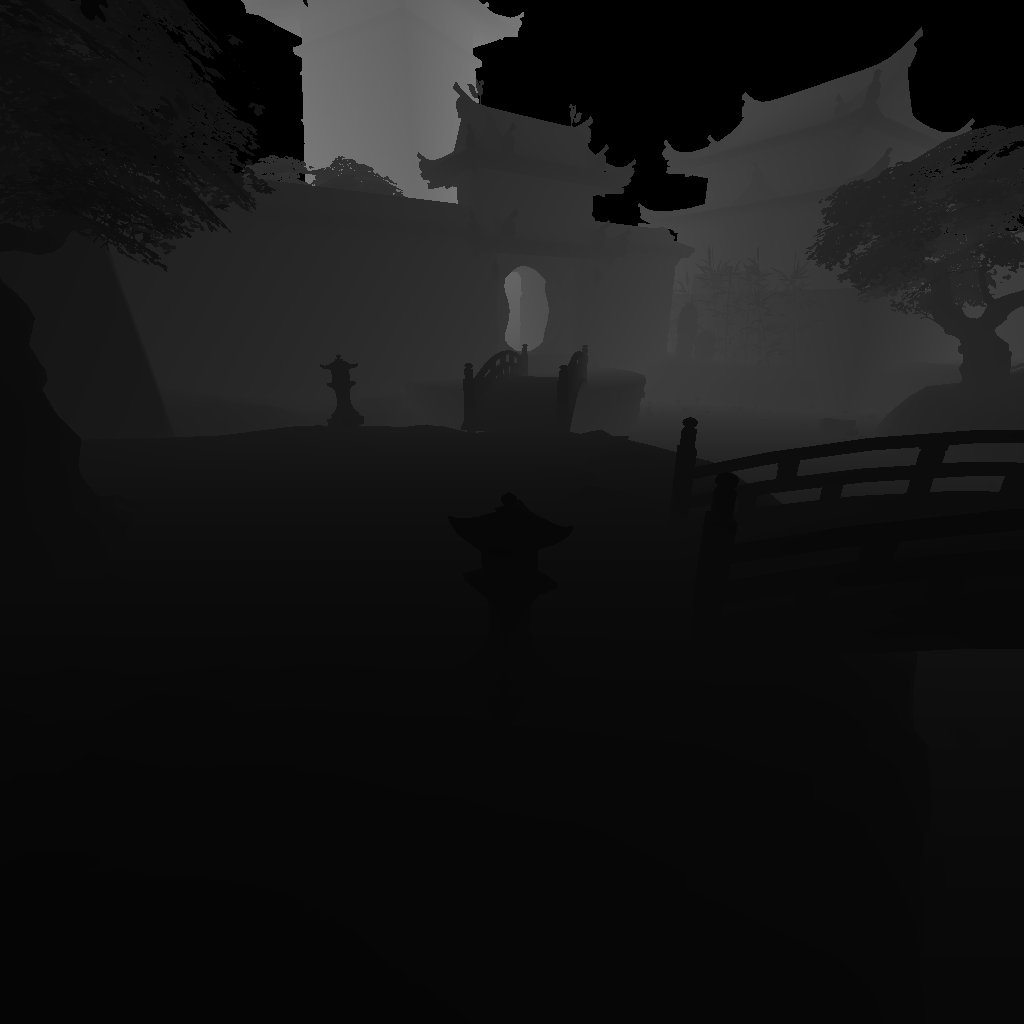}&
    \viditfig{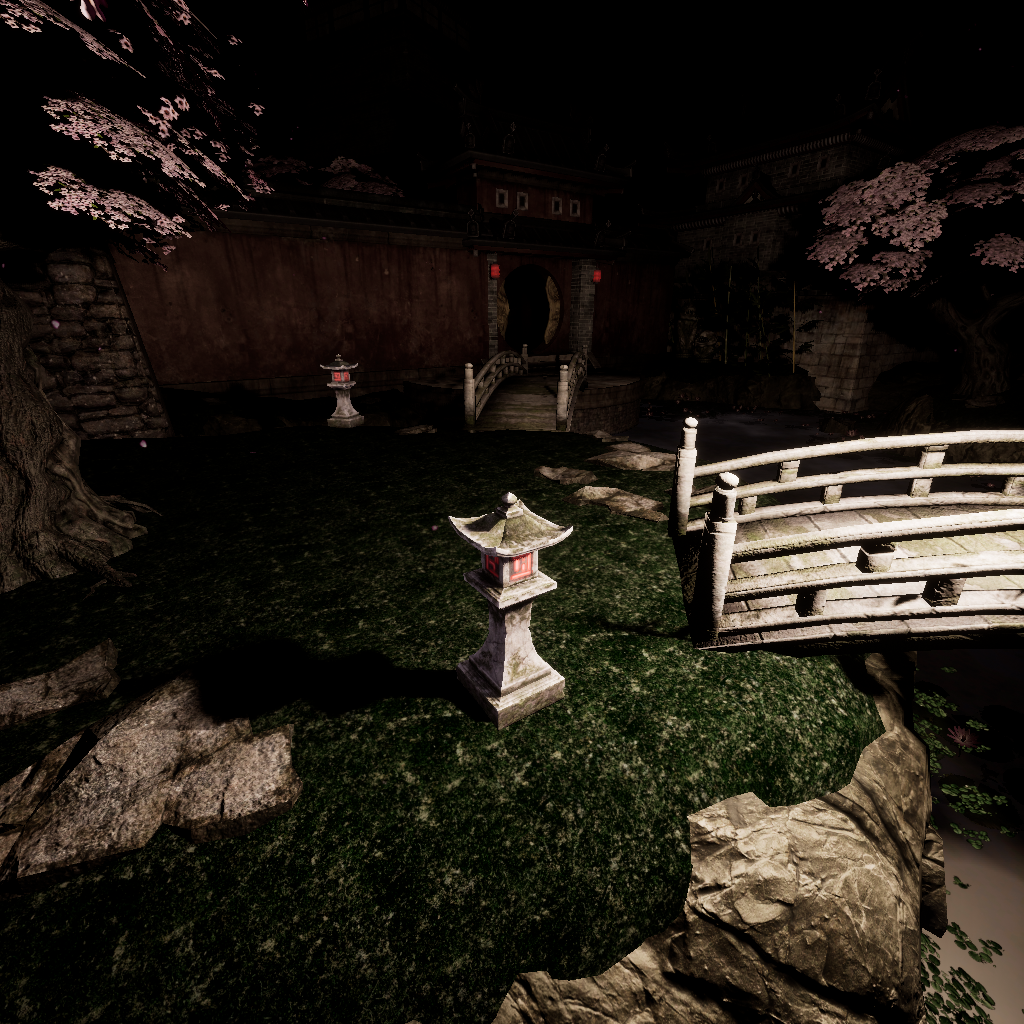} &
    \viditfig{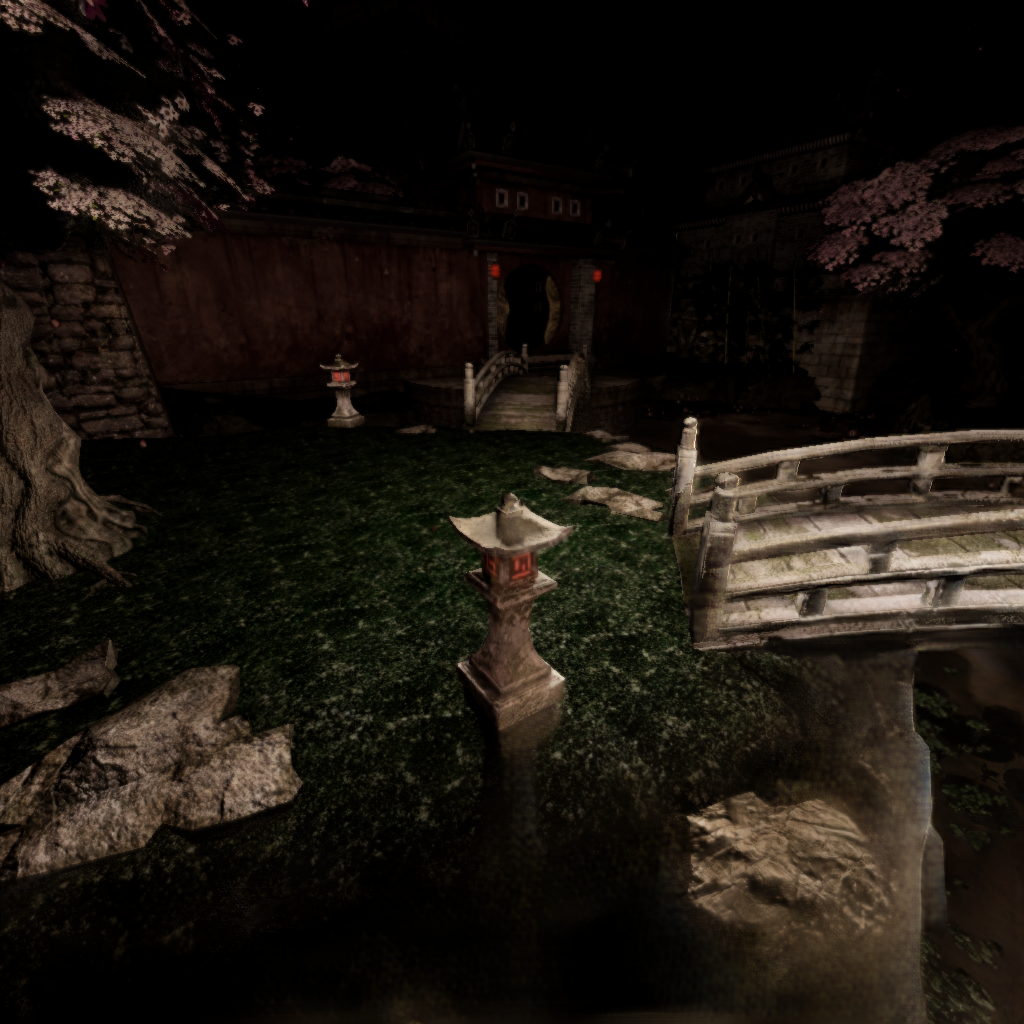} & 
    \viditfig{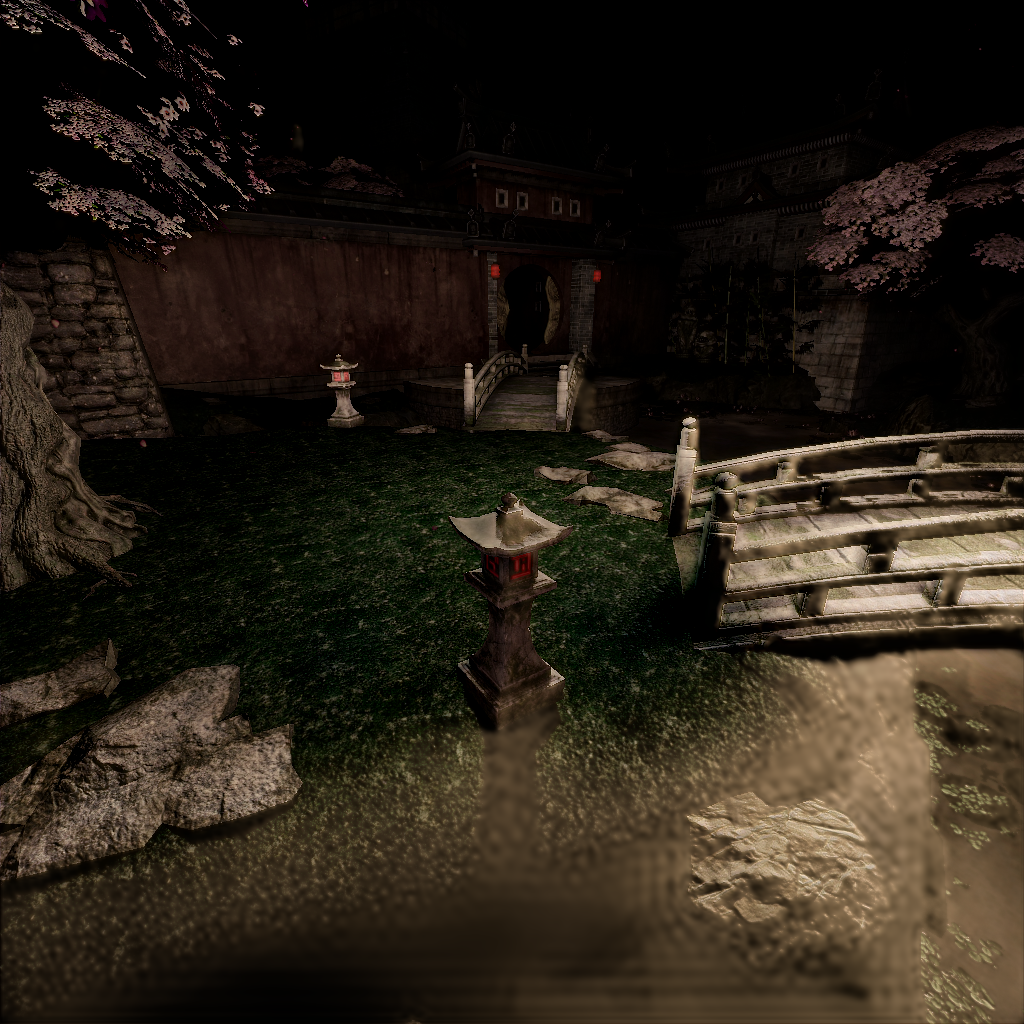} & 
    \viditfig{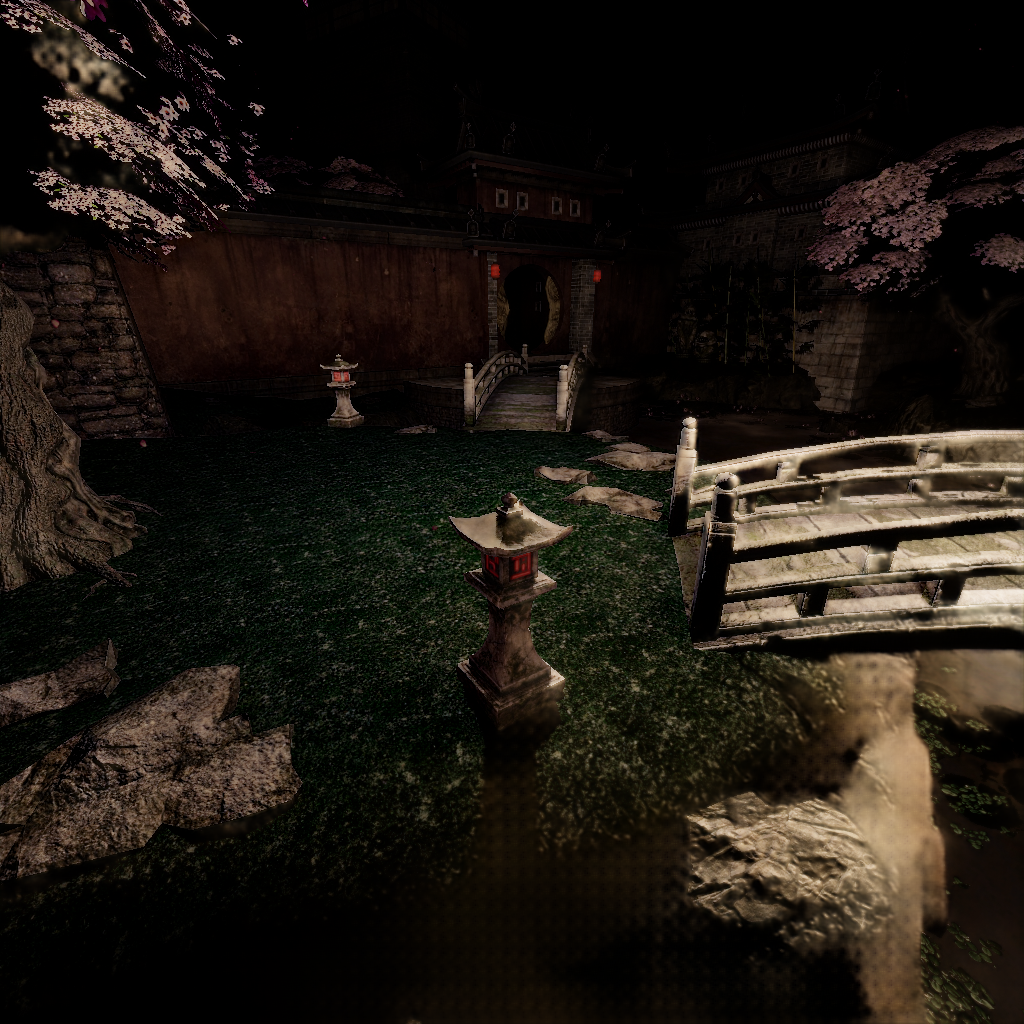} \\
     Input &  Depth &  Ground-truth & AICSNTU-MBNet &  iPAL-RelightNet &  NTUAICS-ADNet\\
    \end{tabu}
\caption{A challenging example image from the NTIRE 2021 Image Relighting Challenge Track 1 (One-to-one relighting) with the output results of some top submission methods.}
\label{fig:track1}
\end{figure*}

\begin{figure*}[htp!]
 \centering
    \begin{tabu}{cccccc}
    \viditfig{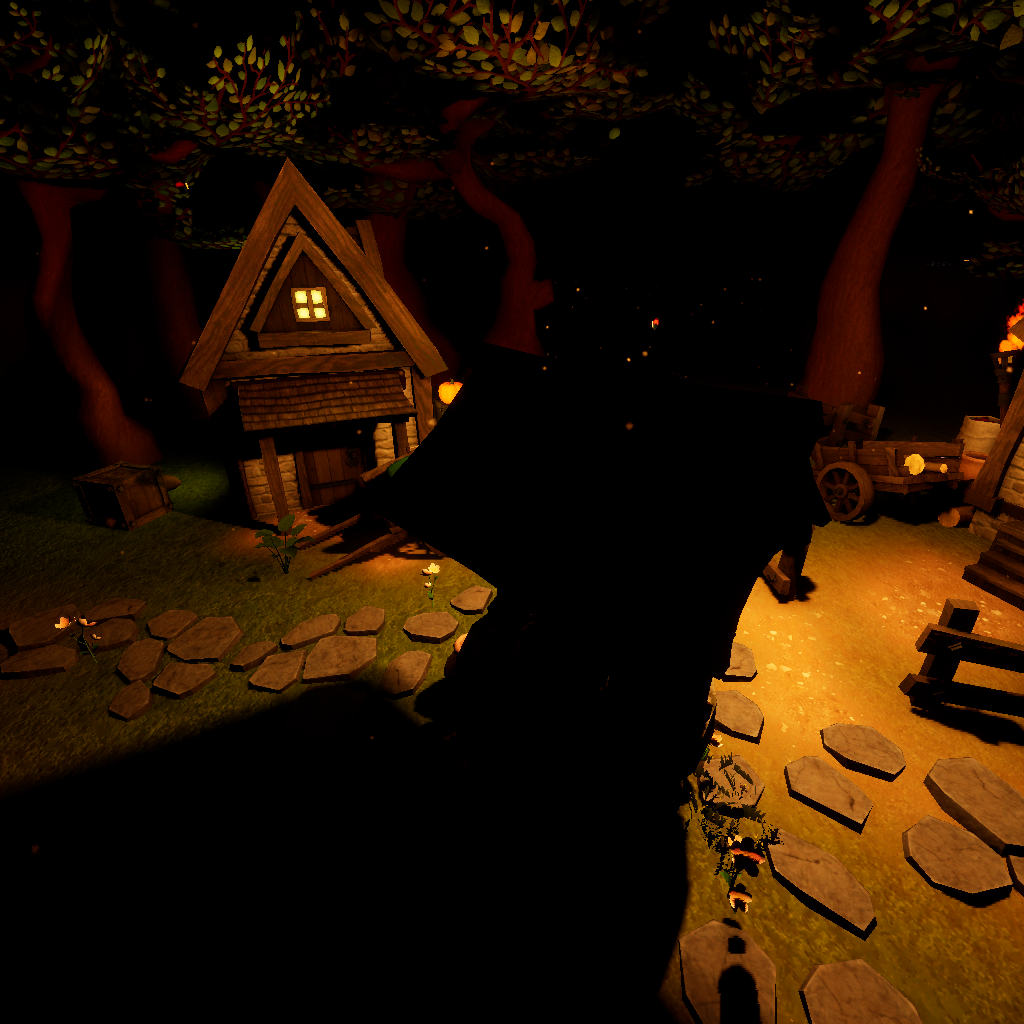} &
    \viditfig{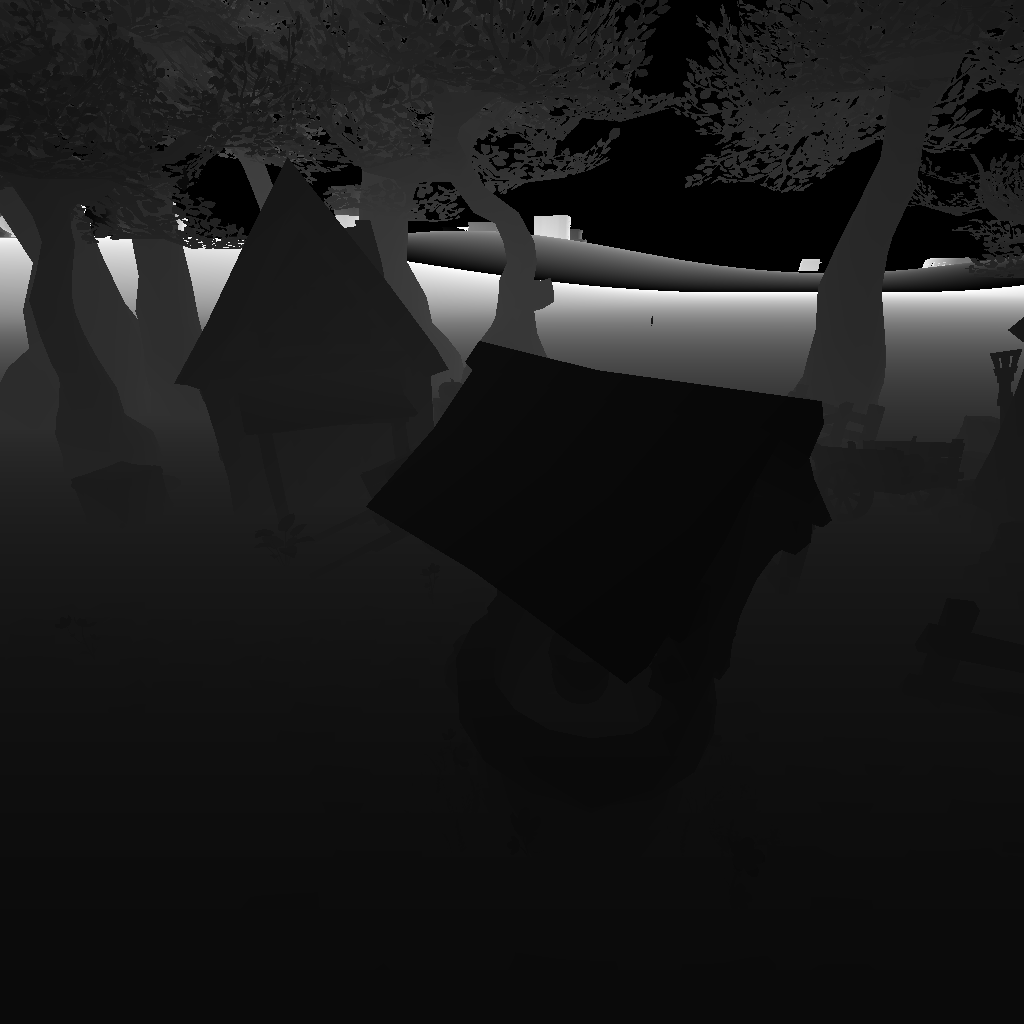} &
    \viditfig{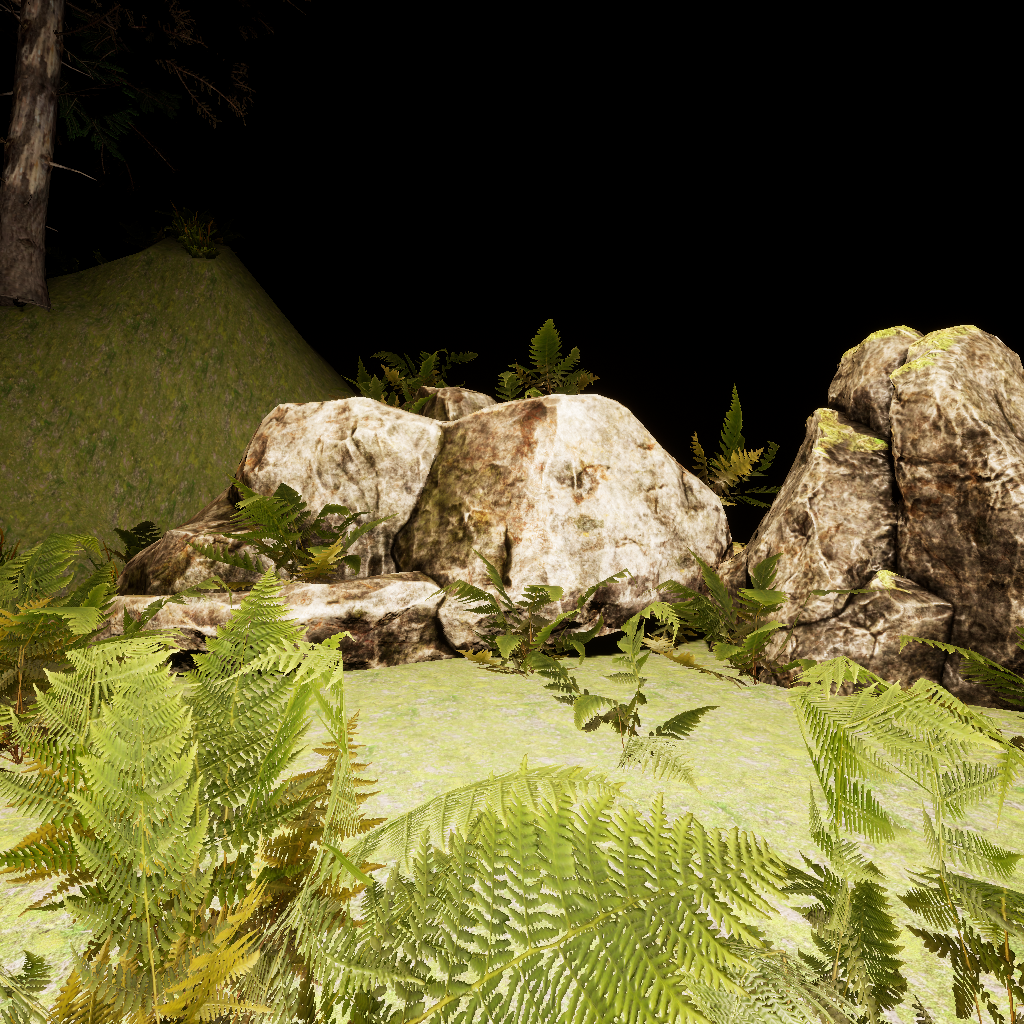} &
    \viditfig{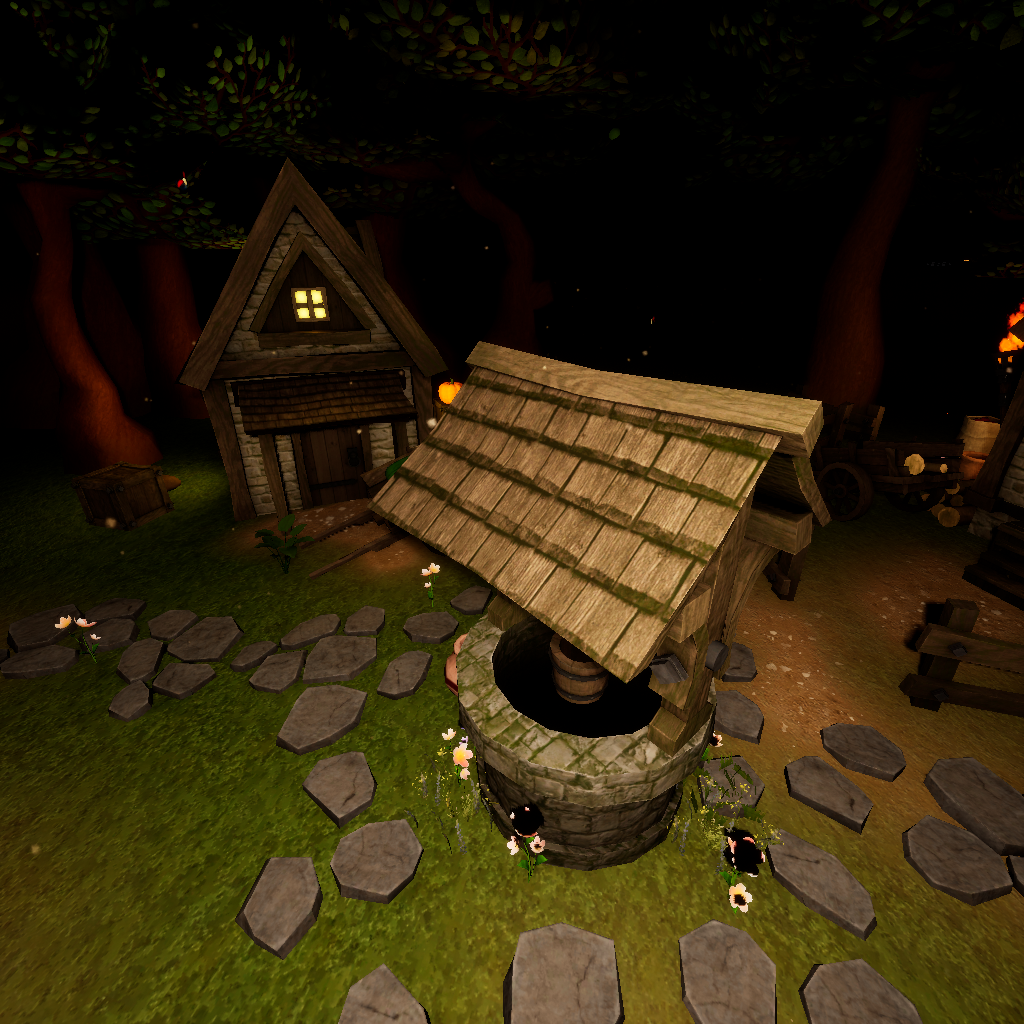} & 
    \viditfig{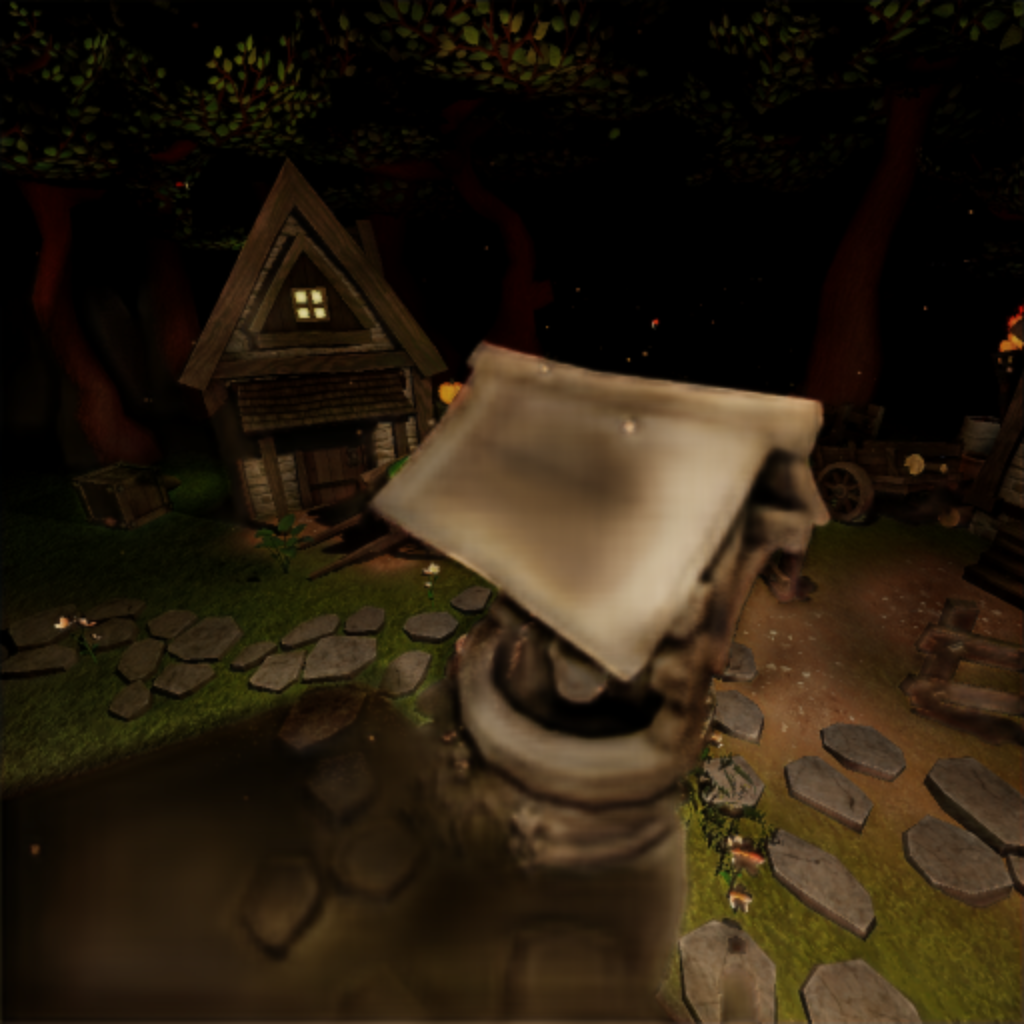} & 
    \viditfig{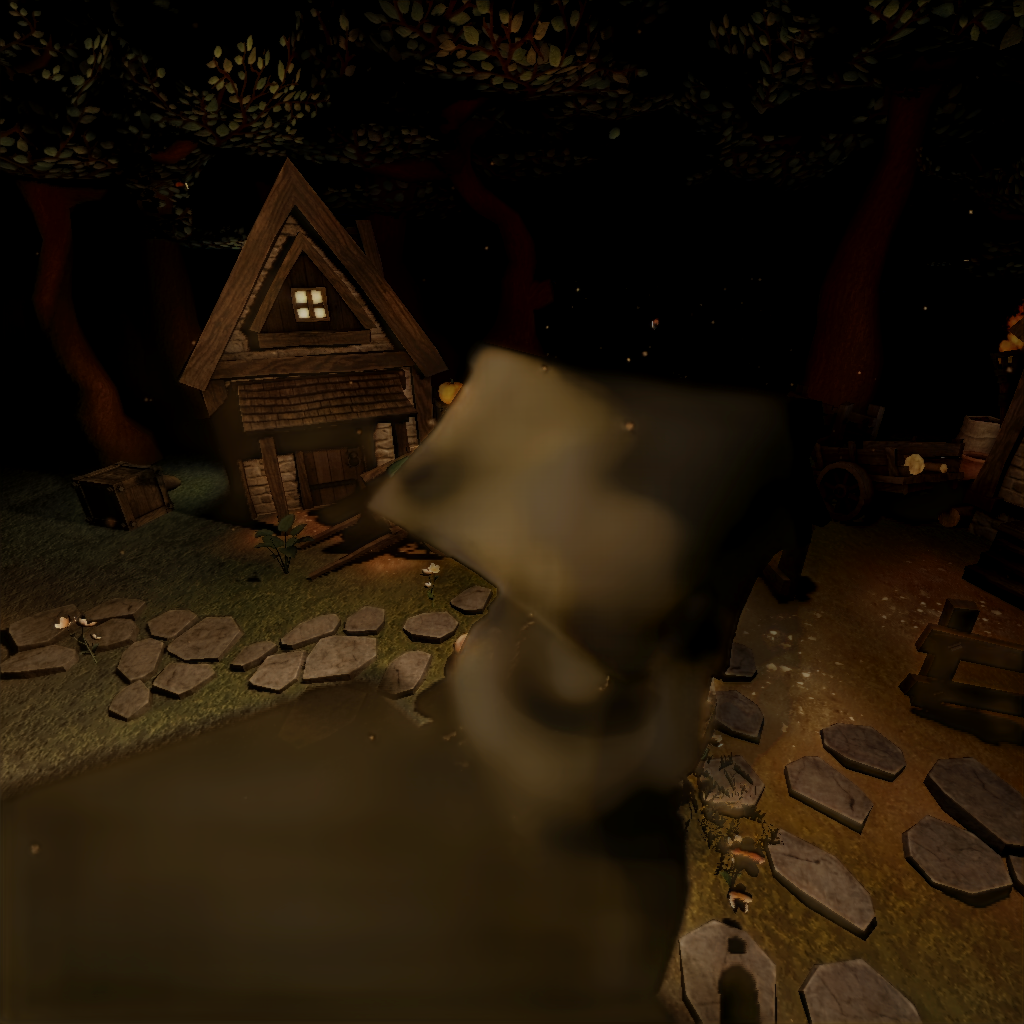}\\
    Input & Depth & Guide & Ground-truth & DeepBlueAI & VUE\\
    %  if 3rd place submission pic is needed, it's IMAGES/T1/3-Pair037.png
    \end{tabu}
\caption{A challenging example image from the NTIRE 2021 Image Relighting Challenge Track 2 (Any-to-any relighting) with the output results of some top submission methods.}
\label{fig:track2}
\end{figure*}

\section{Track 1 methods}
\subsection{AICSNTU-MBNet: Multi-modal Bifurcated Network for depth guided image relighting (MBNet)}
\begin{figure}[ht]
  \centering
\includegraphics[width=0.47\textwidth]{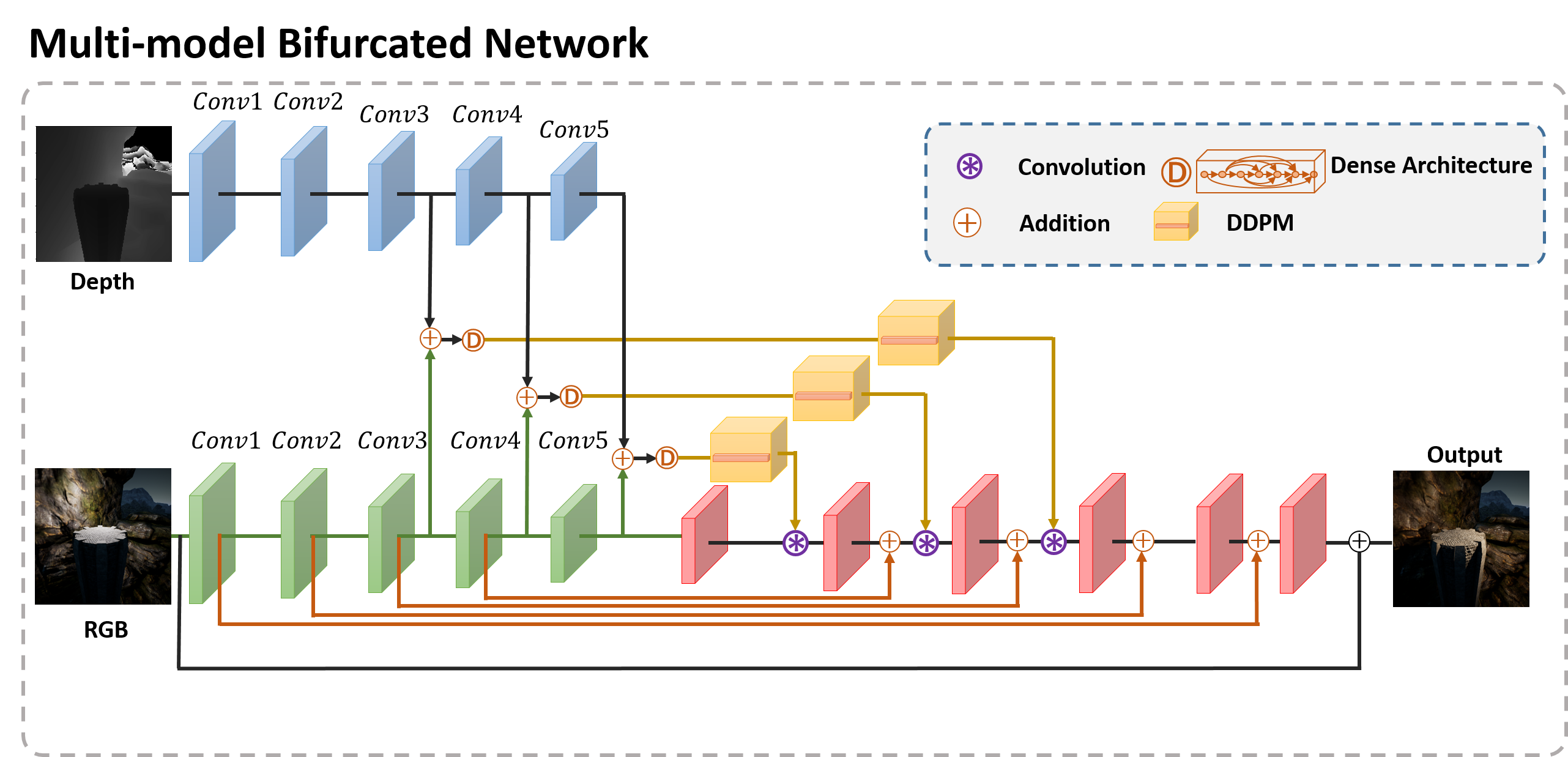}

    \caption{{The architecture of the proposed multi-modal bifurcated network. The network consists of two streams: depth stream and RGB-image stream. We use the dense architecture and DDPM for better feature extraction.
}}
\label{fig:architecture_MB}
\end{figure}

The method details are described in \cite{yang2021multi}. As shown in \figref{fig:architecture_MB}, we rely on the HDFNet~\cite{pang2020hierarchical} to fulfill the depth guided image relighting task. As shown in \figref{fig:architecture}, the proposed network consists of two structures to extract the depth and image features. We apply the two ResNet50 networks as backbones. Three depth and image features from conv3, conv4 and conv5 are fused to achieve representative features. To fuse these two features with multi-receptive fields, we leverage the densely connected architecture to generate the combined features with rich texture and structure information. Then, these features are fed to the dynamic dilated pyramid module (DDPM)\unskip~\cite{pang2020hierarchical} that can generate a more discriminative result. Then, the output of DDPM combines with the output of the decoder by convolving with the multi-scale convolution kernels\unskip~\cite{chen2019pms,ren2016single}. In the decoder part, similar to U-net, we gradually magnify the feature maps and implement a skip connection to concatenate the identical-size feature maps. Furthermore, we make our network learn the residual instead of the full images: the final output is the difference of the original image and the relit image.

\subsection{iPAL-RelightNet: One-to-One Intrinsic Decomposition-Direct RelightNet (OIDDR-Net)}
\begin{figure}[ht]
    \centering
    \includegraphics[width=.48 \textwidth]{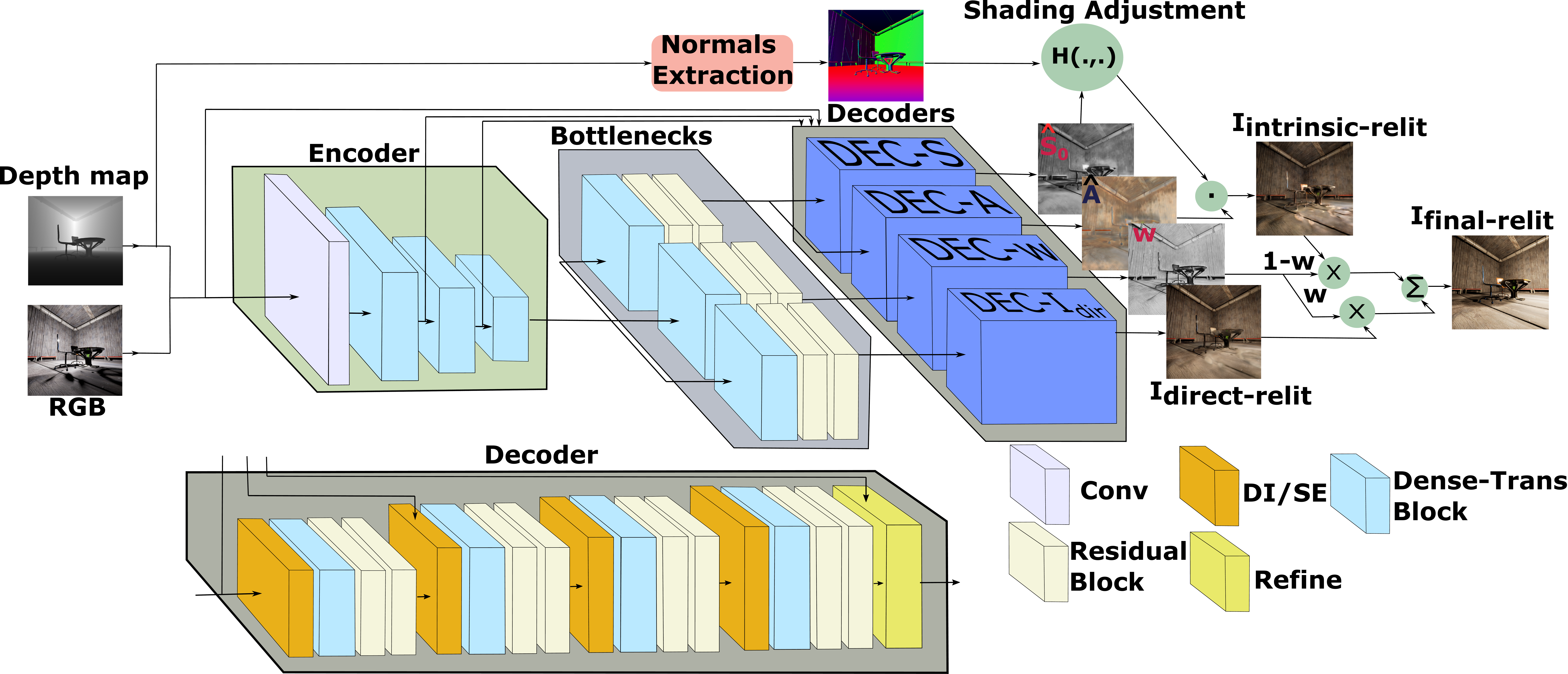}
    \caption{OIDDR-Net architecture.}
    \label{fig:OIDDR}
\end{figure}

This approach \cite{Amir21} exploits two different strategies for generating the relit image with new illumination settings (\figref{fig:OIDDR}). In the first strategy, the relit image is estimated by first predicting the albedo (material reflectance properties) and shading (illumination and geometry properties) of the scene. The intial shading estimation is refined using the normal vectors of the scene which in particular leads to better addition or removal of deep shadows. The rendering rule \cite{rendering} is then used to estimate the relit image ($I_{\textit{intrinsic-relit}}$). In the second strategy, the relit image ($I_{\textit{direct-relit}}$) is generated based on a black box approach, in which the model learns to predict the output based on the ground-truth images and loss terms in the training stage. The two estimates are fused to generate the final relit output ($I_{\textit{final-relit}}$) using a spatial weight map (w), which is learned during the training stage. The source code for the two tracks is made available online\footnote{\url{https://github.com/yazdaniamir38/Depth-guided-Image-Relighting}}. 
% \href{}{\underline{github/Relighting}}

\subsection{MCG-NKU: Pyramidal Depth Guided Relighting Network}
This method handles the relighting task in a coarse-to-fine manner, which has two exclusive 
branches designed for RGB input and depth. 
The depth branch has a cascade structure. It can provide diverse information for different 
scales and attempts to extract high-level information.
The RGB branch has three encoders to tackle the input at different scales and three corresponding decoders that take features from the RGB encoders and the depth branch as input and generates the relit images.
We find that the surface normal is an important prior that induces the network to learn local illumination. Additionally, we utilize an explicit approach with intuition borrowed from 
Transformer networks. 
Specifically, we emphasize the relative positional information by using linear positional encoding that encodes the x-axis/y-axis relative position through two feature maps.
\begin{figure}[t]
	\centering{\includegraphics[width=\linewidth]{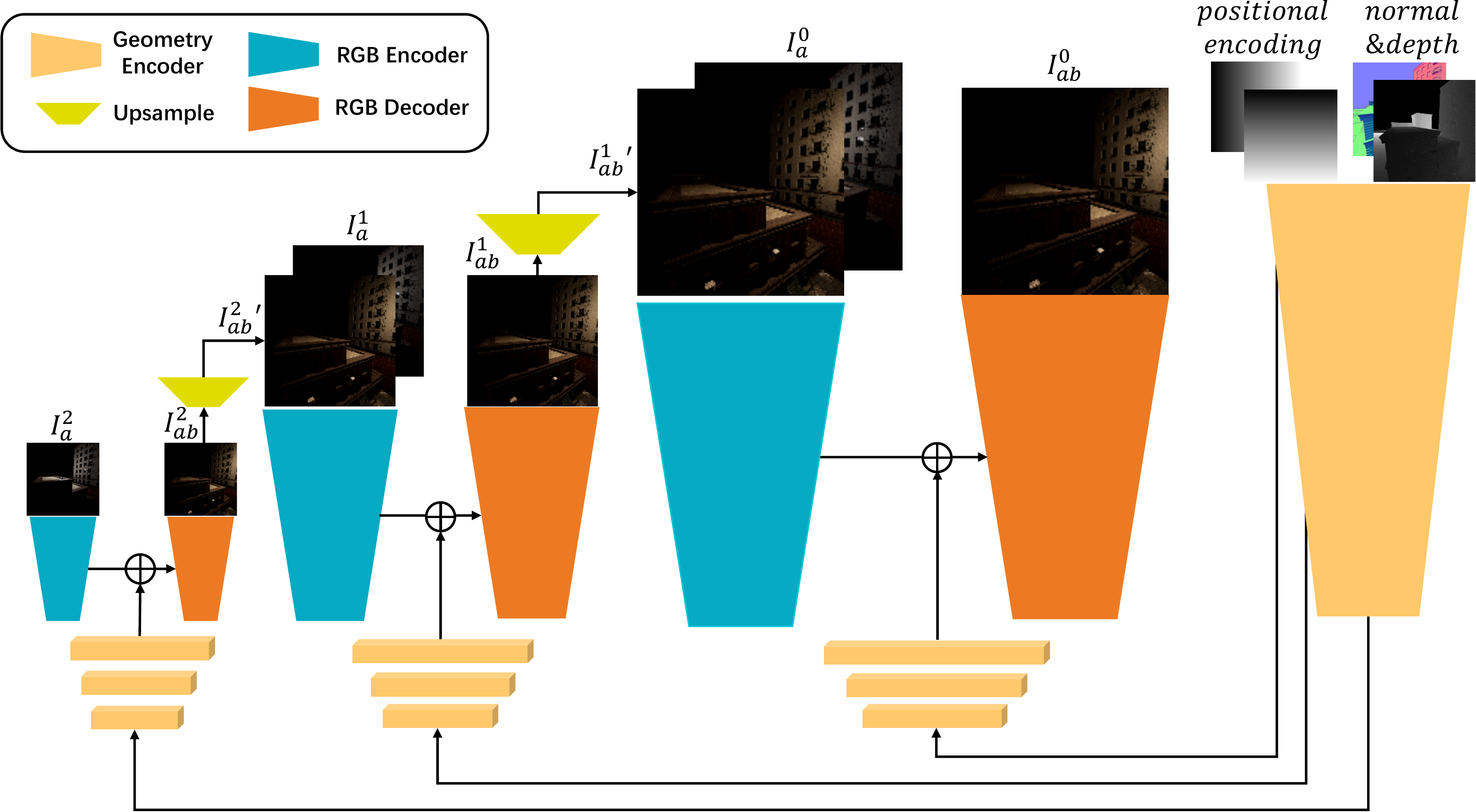}}
	\caption{Overview of the proposed pyramidal depth guided relighting network.}
	\label{fig:PDGRN}
\end{figure}
The overall structure of our proposed method is illustrated in Fig.~\ref{fig:PDGRN}.

% team vue
\subsection{VUE: Deep RGB and Frequency Domain CNNs for Image Relighting (DRFNet)}
\begin{figure}[t]
	\centering{\includegraphics[width=\linewidth]{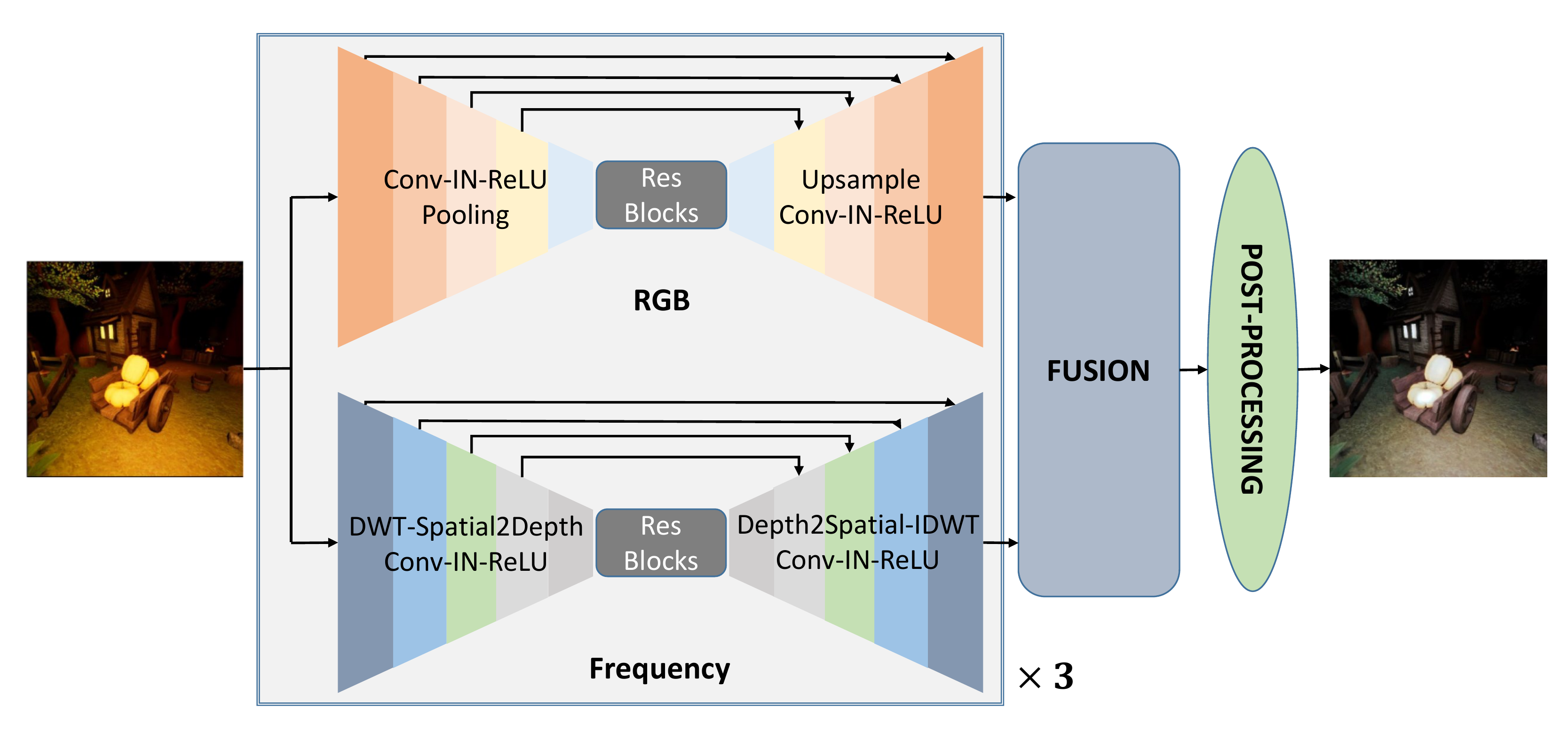}}
	\caption{Overall architecture of the DRFNet in track 1.}
	\label{fig:vue_relighting-1}
\end{figure}

We make use of both the RGB domain \cite{10.1007/978-3-030-67070-2_32} and the frequency domain~\cite{puthussery2020wdrn,elhelou2020stochastic,zhou2019comparative} CNNs for one-to-one image relighting by end-to-end training, and the overall pipeline of our solution is illustrated in Fig.~\ref{fig:vue_relighting-1}.
Intuitively, the frequency domain and the RGB domain of images are two distinctive domains. Building mappings from one lighting setup to the other illumination settings in the RGB space and frequency space could be complementary.
These CNNs are all designed as U-Net encoder-decoder architectures with skip connections. 
Specifically, for RGB CNNs, the encoder is composed of N ``Conv-IN-ReLU-Pooling" blocks and its decoder contains N ``Upsample-Conv-IN-ReLU" building blocks; for frequency domain CNNs, the corresponding building blocks in its encoder and decoder are ``DWT-Spatial2Depth-Conv-IN-ReLU" and ``Depth2Spatial-IDWT-Conv-IN-ReLU", respectively.
There are also M ResBlocks between the encoder and decoder. The fusion module simply performs per-pixel averaging. 
As validated in the developing phase of this challenge, slightly lowering the luminance and enhancing the color saturation are effective post-processing strategies. 
Note that depth images are provided, thus in our implementation the CNNs take as input the concatenation of the RGB image and its depth image in both training and testing phases.
% vue end

% team DeepBlueAI
\subsection{DeepBlueAI: Deep Fusion Network for Image Relighting (DFNIR)}
The team uses the method and model weights trained in track 2 and simply selects one image with the pre-defined illumination settings (East, 4500K) from the training set as the guide image to get the results for track 1. More details can be found in the description of the track 2 solution~\ref{DFNIR}.

\subsection{NTUAICS-ADNet: Depth Guided Image Relighting by Asymmetric Dual-stream Network}
\begin{figure}[ht]
  \centering
\includegraphics[width=0.5\textwidth]{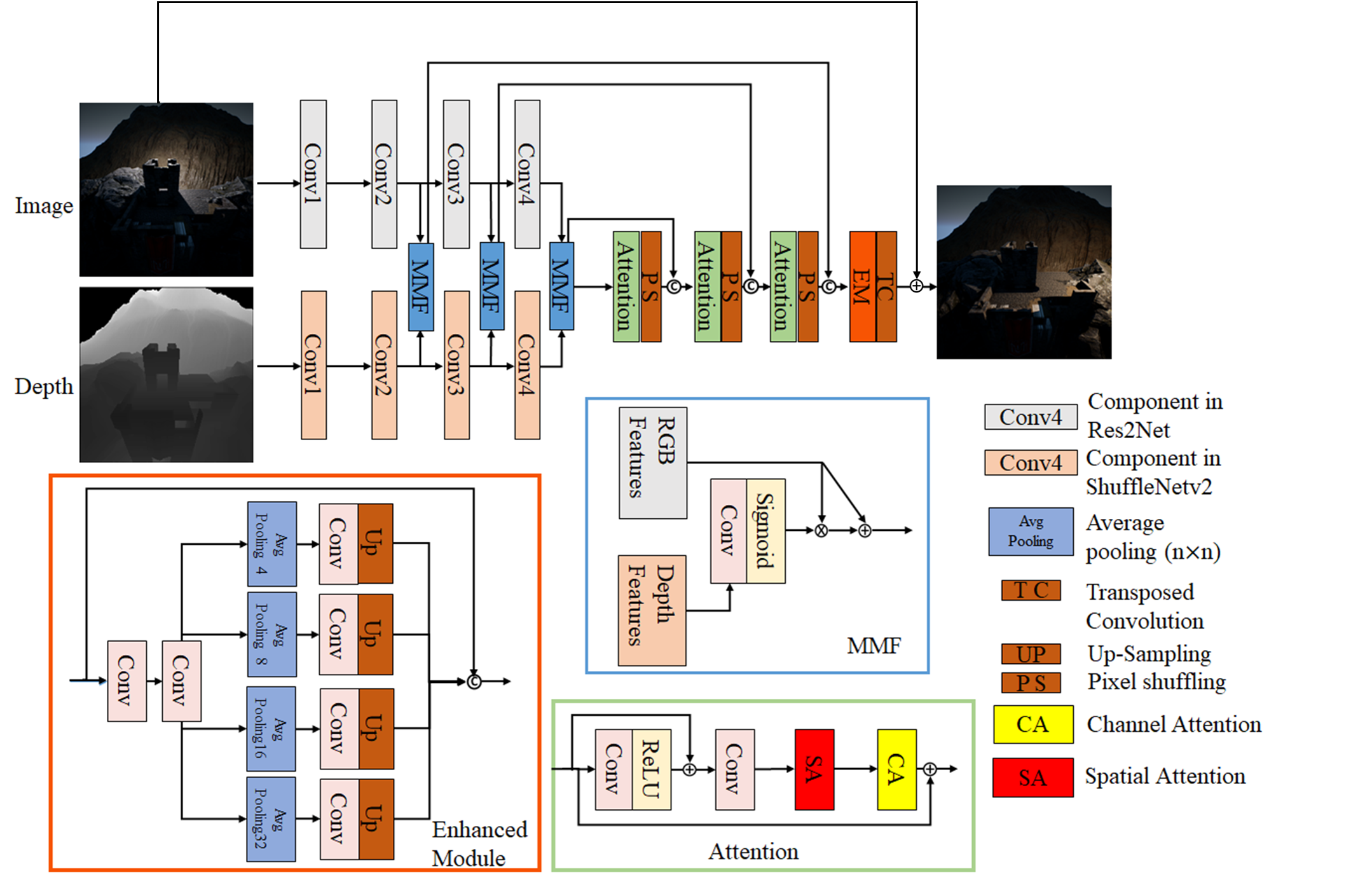}

    \caption{{The proposed ADNet. This model applies the Res2Net and ShuffleNetv2 as the encoder to extract the image and depth map feature representations. We design the Multi-Modal Fusion block to integrate these two feature representations. In the decoder parts, we use the attention mechanism and the enhanced module to refine features and relight the image.
\label{fig:architecture}
}}

\label{fig:figure2}
%\vspace{-0.5cm}
\end{figure}

The overview of our solution ADNet is illustrated in \figref{fig:architecture}. As shown in \figref{fig:architecture}, we first use a powerful network to extract the RGB-image feature representation. Both the encoder and the decoder are based on the Res2net~\cite{gao2019res2net} network and the decoder contains an attention mechanism (Attention)\unskip~\cite{chen2020pmhld} and enhanced modules (EM), which is motivated by \cite{qu2019enhanced}. The details of the EM and Attention approaches are demonstrated in \figref{fig:architecture} (the orange box and the blue box). The attention block consists of both spatial and channel attention mechanisms. For the feature extraction of the depth map, we utilize the smaller backbone, that is, the ShuffleNet V2 \cite{ma2018shufflenet} to extract the depth map feature representation. Moreover, we combine the features in Conv 2, Conv 3, and Conv 4 with the multi-modal fusion (MMF) module. This module contains a convolutional layer to make the size of the depth feature representation identical to that of the RGB-image feature representation. And these two feature representations are multiplied and summarised to become refined features. The latter features are concatenated and combined with the representation from the decoder features. It is noted that the target output of the network is designed so as to learn the residual components between the output and input instead of the whole image.

\subsection{NTUAICS-VGG: Two Birds, One Stone: VGG-based Network with VGG-based Loss for Depth Guided Image Relighting}
\begin{figure}[ht]
  \centering
\includegraphics[width=0.5\textwidth]{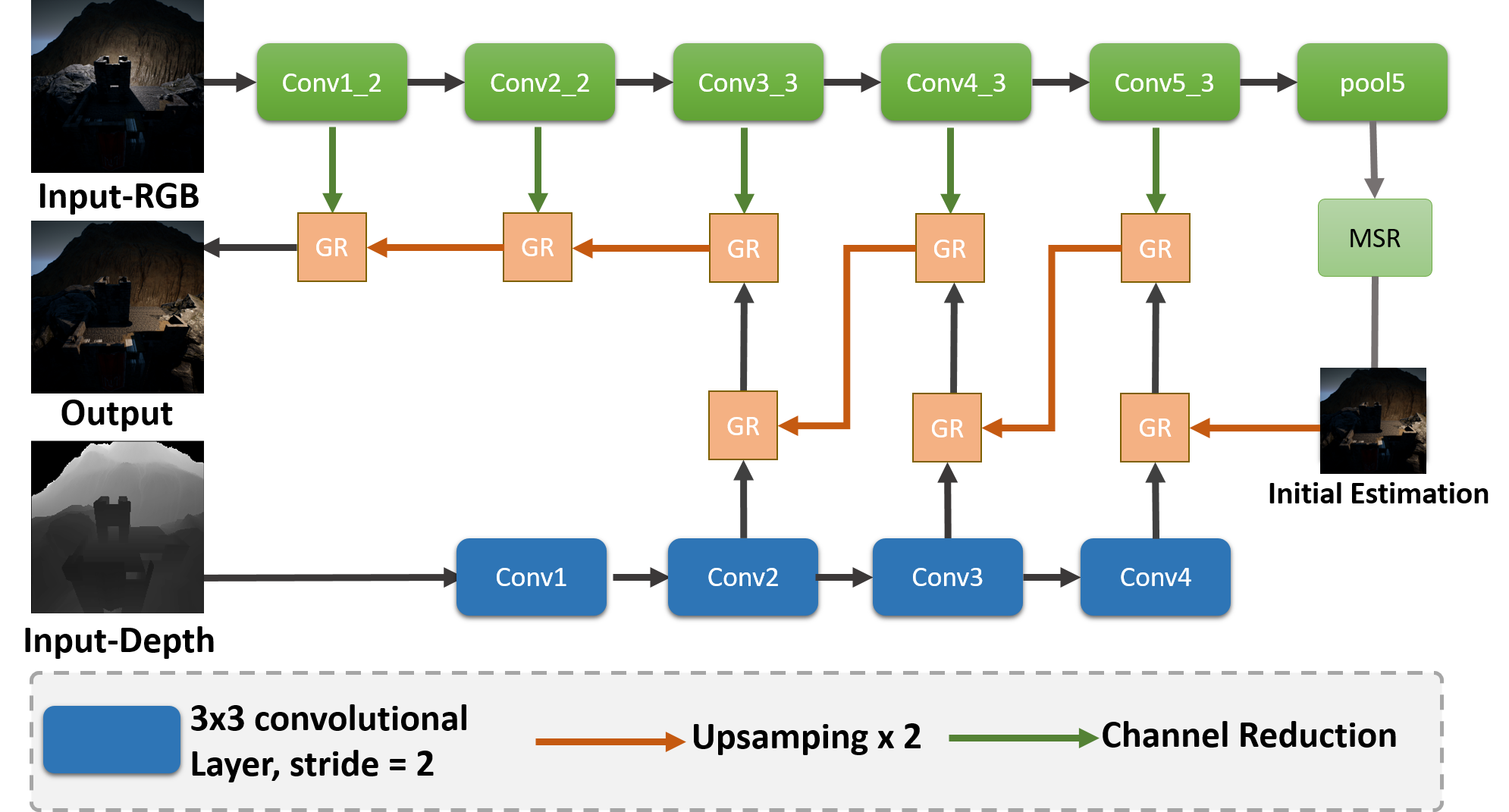}

    \caption{{The architecture of the proposed relighting network. The features of RGB and depth information are fed into GR blocks for generating the final output.
}}
\label{fig:VGG1}
%\vspace{-0.5cm}
\end{figure}

Inspired by\unskip~\cite{chen2020progressively}, we design two branches to extract the RGB-image feature representation and the depth feature representation, respectively. For the image features, we apply the VGG-16 to extract multi-scale representations. Then, these features are fed into the multi-scale residual block (MSR) to generate the initial estimation of the relit image. The architecture of MSR is presented in \figref{fig:VGG1}. Then, the initial estimation of the relit image is fed into a series of guided residual blocks (GR)\unskip~\cite{chen2020progressively} for further refinement. The GR blocks use the features extracted by each layer in the RGB branch and depth branch to reconstruct the details of the final output. For the depth feature representation, we simply apply a single layer of convolution ($3\times3$ convolutional kernels with 64 channels) to extract three kinds of features. Then, with the feature extraction of the two branches and the refinement process, the final relit output is generated.

We rely on the $\ell 1$ Chabonnier loss \cite{barron2019general}, the Wavelet SSIM loss \cite{yang2020net}, and the visual perceptual loss \cite{johnson2016perceptual} to train our model. We use the Adam optimizer with a learning rate of 0.0001 and the learning rate decreases by 0.1 every 50 epochs. The total number of training epochs is 200 and the training takes 6 hours when setting the batch size to 3 and using V100 GPUs.

\subsection{Wit-AI-lab: Multi-scale Self-calibrated Network for Image Light Source Transfer (MCN)}
\begin{figure}[t]
	\centering{\includegraphics[width=\linewidth]{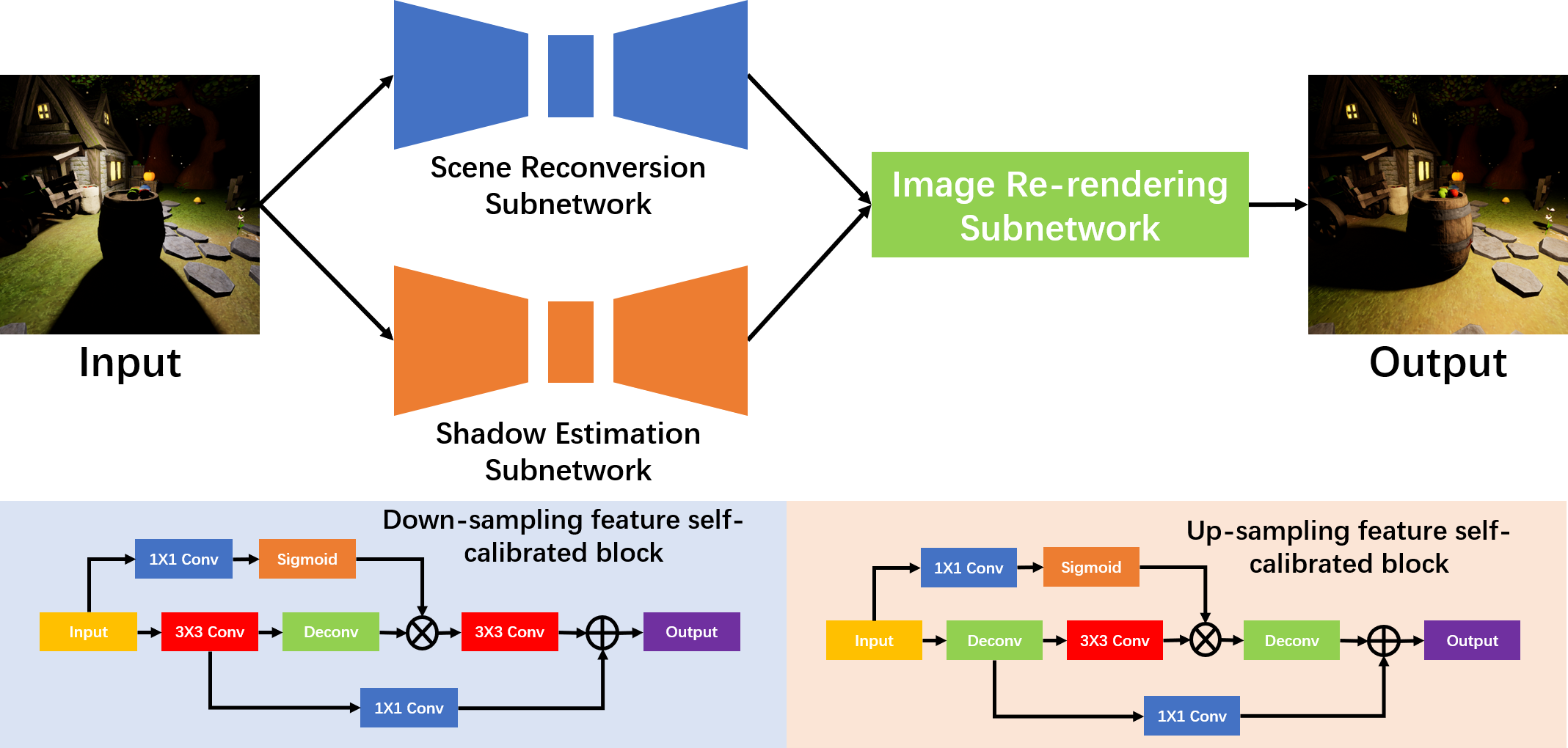}}
	\caption{Overall architecture of the MCN.}
	\label{fig:MCN_Network}
\end{figure}

Inspired by \cite{DRN}, the architecture of MCN is shown in Fig.~\ref{fig:MCN_Network}, which consists of three parts: scene reconversion subnetwork, shadow estimation subnetwork, and image re-rendering subnetwork.
First, the input image is processed in the scene reconversion subnetwork to extract primary scene structures. At the same time, the shadow estimation subnetwork aims to the change of the lighting effect. Finally, the image re-rendering subnetwork learns the target color temperature and reconstructs the image with primary scene structure information and predicted shadows. 
Considering that image light source transfer is a task of recalibrating the light source settings, the team proposes a novel downsampling feature self-calibrated block (DFSB) and upsampling feature self-calibrated block (UFSB) as the basic blocks for scene reconversion and shadow estimation tasks to calibrate feature information iteratively, thereby improving the performance of light source transfer. 
In addition, the team designs the multi-scale feature fusion method for the scene reconversion task, which provides more accurate primary scene structure for the image re-rendering task.

%Team: alphaRelighting
\subsection{alphaRelighting: Light Transfer Network for Depth Guided Image Relighting (LTNet)}

\begin{figure}[h!]
	\centering{\includegraphics[width=\linewidth]{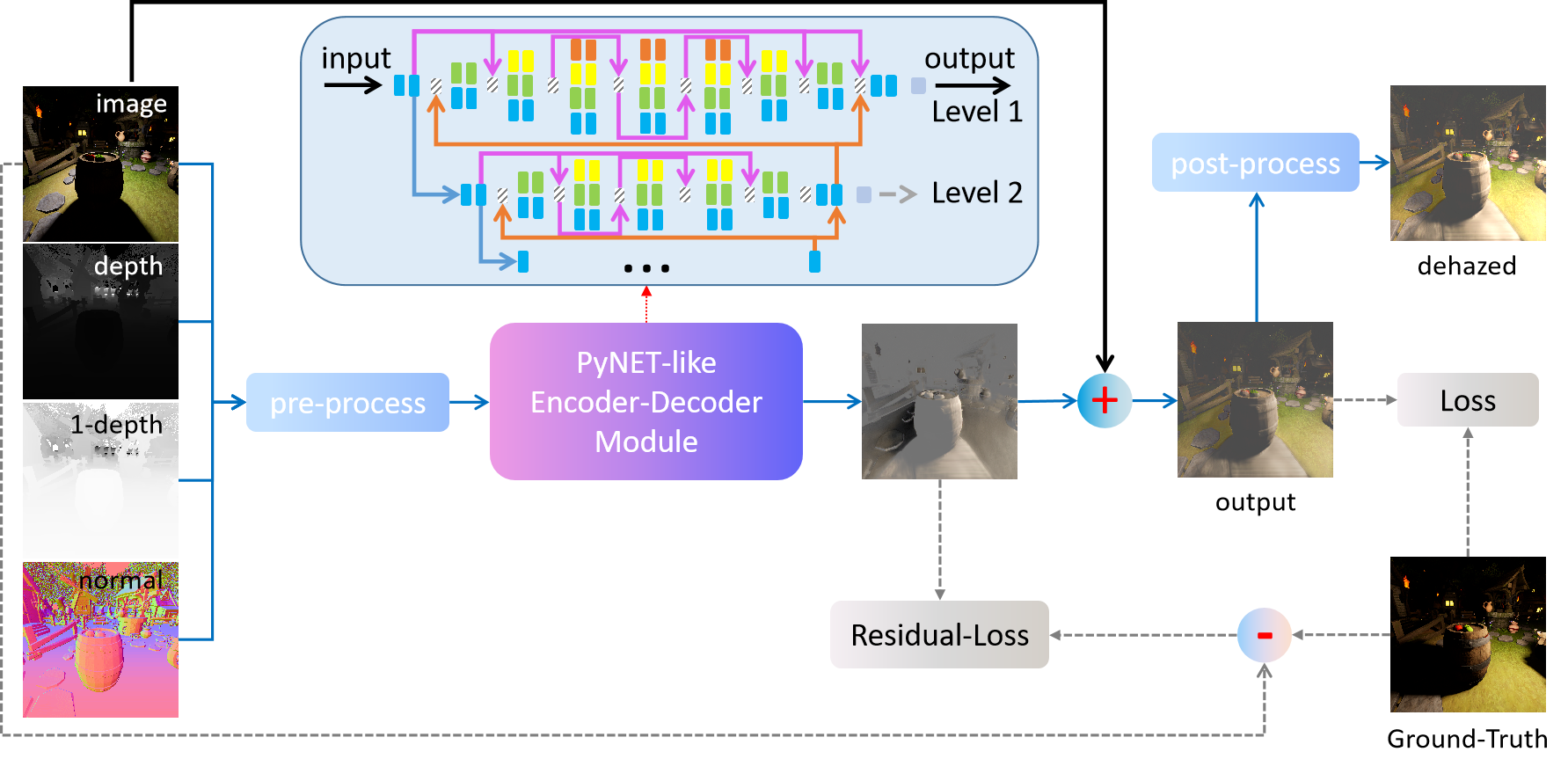}}
	\caption{Overall architecture of the LTNet.}
	\label{fig:LTNet}
\end{figure}

As shown in Fig.~\ref{fig:LTNet}, the key modules of the proposed LTNet~\cite{zhu2021ltnet} are the global skip connection, the PyNET-like encoder-decoder network and the additional residual loss $L_{RES}$. The global skip connection enables the input to be added to the output of the encoder-decoder network directly. Then, the encoder-decoder network is set to learn the residual between the ground-truth image ($I_{GT}$) and the input image ($I_{input}$). Besides, we add a residual ground-truth ($I_{GT}-I_{input}$) to supervise the training as well as the original loss supervised by $I_{GT}$. LTNet uses a PyNET-like encoder-decoder network~\cite{PyNET} to learn the residual of the ground-truth image and the input image, which adopts a slightly dense connection and a number of convolution blocks in parallel with convolution filters of different sizes. In LTNet, the global skip connection plays a key role because it forces the network to encode the variance of the input image and the target.

% \subsection{Couger AI: Multi-Level Attention based Efficient U-Net Encoder Decoder Model: A novel approach for One-to-One Image Relighting}

% \begin{figure}[h!]
% 	\centering{\includegraphics[width=\linewidth]{IMAGES/one_one_image.jpg}}
% 	\caption{Proposed Model: A Multi-Level Attention based Efficient U-Net Encoder Decoder Model}
% 	\label{fig:MLAEUEDM}
% \end{figure}

% The present approach for tackling the one-to-one relighting problem is a multi-level U net encoder-decoder\cite{eyeNet} model which consists of i) deep layers of the Efficient-Net B4 \cite{tan2019efficientnet} in the encoder, residual dense blocks along with CBAM\cite{cbam} attention in bottleneck layer, and res2net \cite{res2net} block in the decoder. The Efficient-Net B4 layers used in the encoder are initialized by random weights. However, imbalanced U-Net sometimes causes loss in either very low level features or very high level features, which results in the ambiguous and low level of edges, colors, and contrast of the generated images. We dealt with this problem with multi-level supervision with prior-based high frequency and low frequency components. The detailed architecture is presented in Fig.~\ref{fig:MLAEUEDM}. In the following subsection, we will discuss the details of each part of the proposed architecture.

\section{Track 2 methods}
\subsection{DeepBlueAI: Deep Fusion Network for Image Relighting (DFNIR)}\label{DFNIR}

\begin{figure}[t]
	\centering{\includegraphics[width=\linewidth]{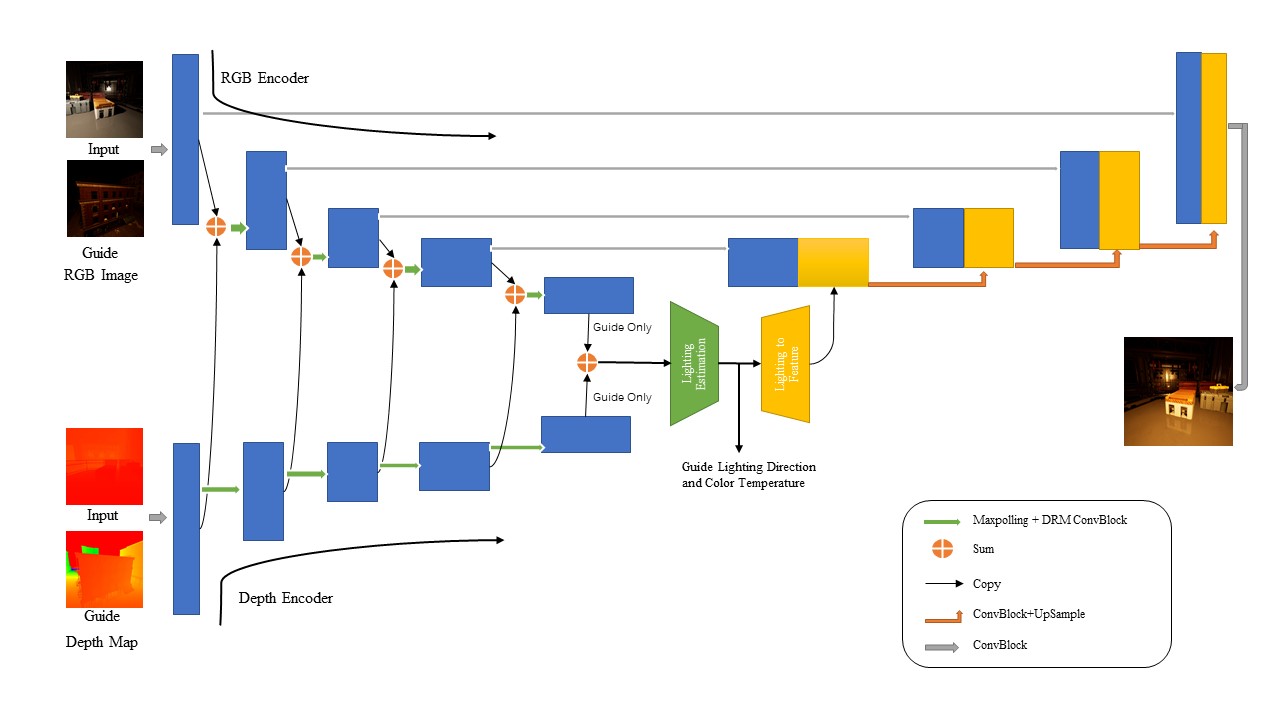}}
	\caption{Overall architecture of the DFNIR.}
	\label{fig:DFNIR_Network}
\end{figure}

The team designed a U-Net style encoder-decoder structure with RGB-D image maps as input and directly outputs the relit RGB image, as shown in Fig.~\ref{fig:DFNIR_Network}. 
In the encoder network, inspired by \cite{fusenet2016accv}, a fusion network is designed to extract feature representations of the input and guide images. To better recognize the illumination of the input and guide images, a Dilated Residual Module (DRM) is designed and used at each level, with three convolutional layers with dilation of 1,3, and 5. Based on successful previous work\cite{10.1007/978-3-030-67070-2_32}, the team also designed and added an illumination-estimator network and an illumination-to-feature network. The difference is that in our solution we completely replace the features of the input image with the features of the guide image output obtained from the illumination-to-feature net. In addition, we use in the skip connections and the decoder network a Residual block with two convolutional layers.

% team vue
\subsection{VUE: Deep RGB and Frequency Domain CNNs for Image Relighting (DRFNet)}
\begin{figure}[t]
	\centering{\includegraphics[width=\linewidth]{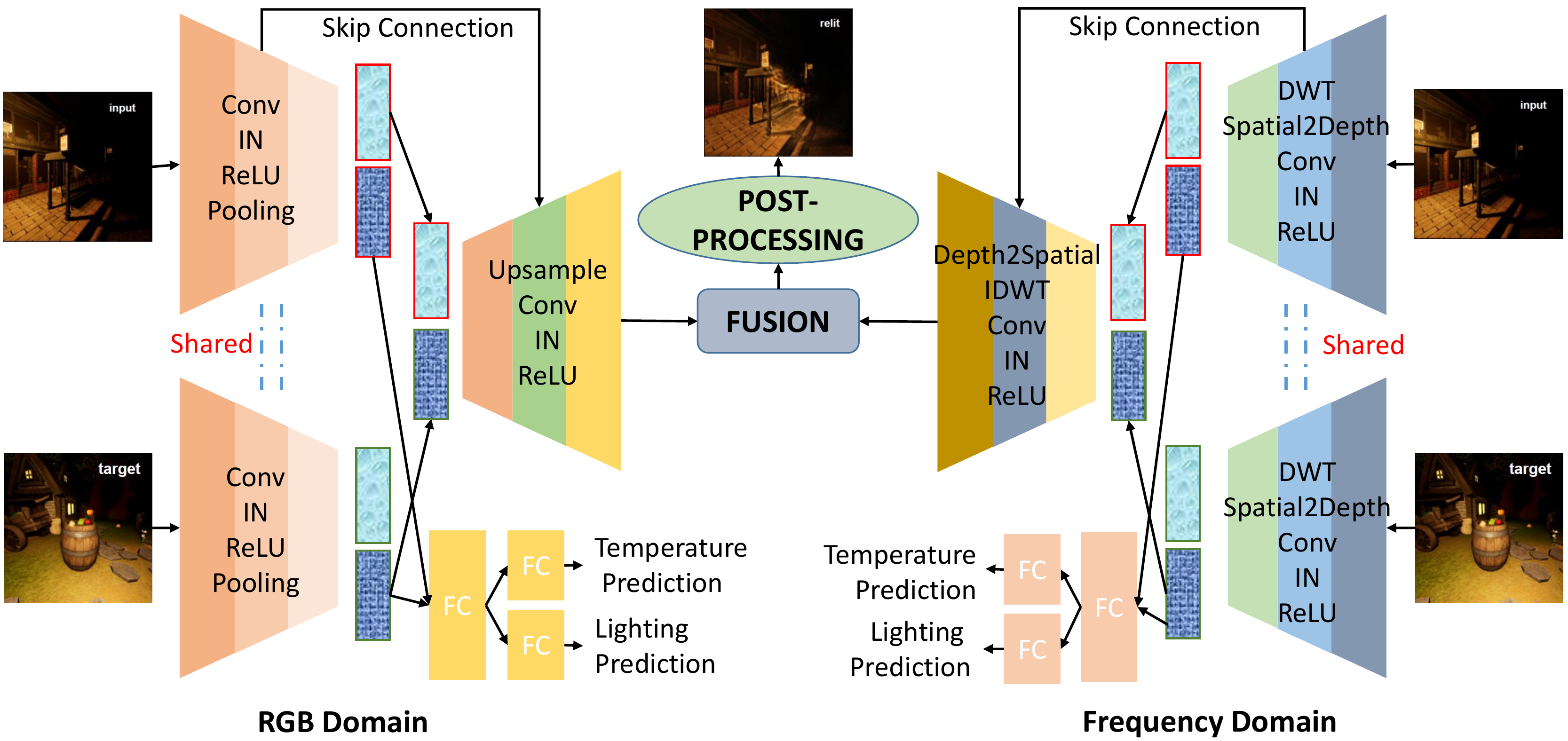}}
	\caption{Overall architecture of the DRFNet in track 2.}
	\label{fig:vue_relighting-2}
\end{figure}

We rely on the same network architecture as in track 1. We utilize both the RGB domain and the frequency domain CNNs for any-to-any image relighting by end-to-end training, and the overall pipeline of our solution is illustrated in Fig.~\ref{fig:vue_relighting-2}.
The difference is that we take both input image and guide image as the inputs of our network.
To support any-to-any mapping, we explicitly split feature maps generated by the encoder into ``content feature" and ``lighting feature". The ``lighting feature" is trained with extra supervision from the temperature prediction and illumination-direction prediction branches. Then, the learned lighting feature of the guide image and the content feature of the input image are combined for decoding the desired relighting output. 
The fusion module simply performs per-pixel averaging. 
% vue end

\subsection{AICSNTU-SSS: A Single Stream structure for depth guided image relighting (S3Net)}
\begin{figure}[ht]
  \centering
\includegraphics[width=0.5\textwidth ]{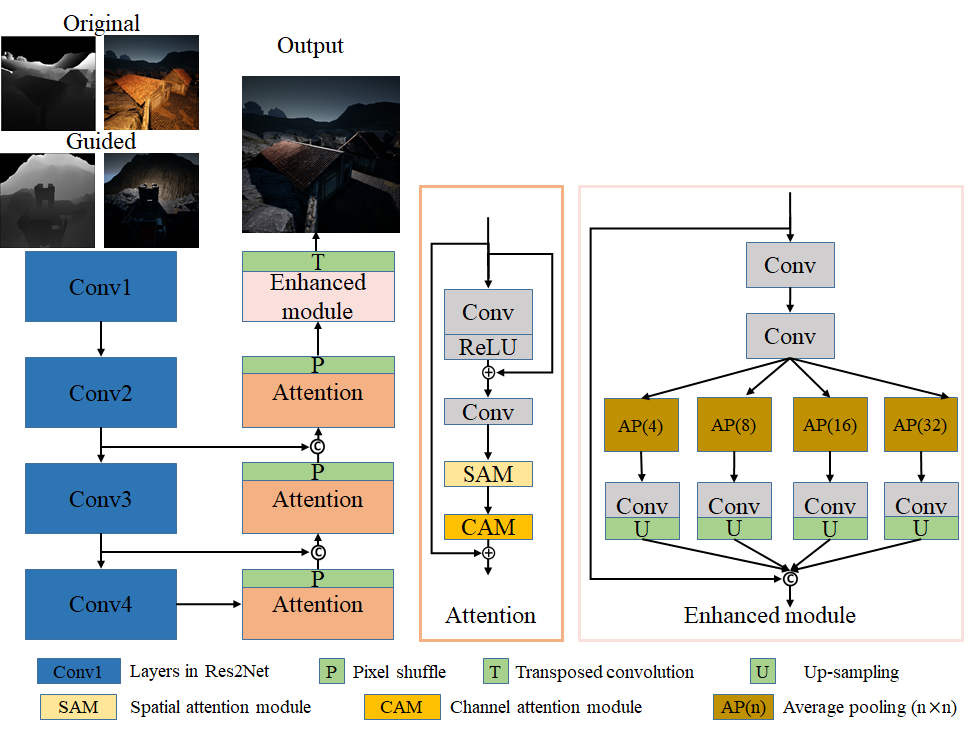}

    \caption{{The proposed network for depth guided image relighting. This model applies the Res2Net and an encoder to extract both the image and depth map features. All images and depth maps are concatenated as input. In the decoder parts, we use an attention mechanism including channel and spatial attention and the enhanced module to refine features and relight the image.
\label{fig:architecture_SSS}
}}

% \label{fig:figure2}
%\vspace{-0.5cm}
\end{figure}

The method directly combines the original image, original depth map, guide image, and guide depth map as the input. This input is seen as the 8-channel tensor and the output is the 3-channel image, refered to as the relit image. We rely on the Res2Net101 \cite{gao2019res2net} network for our backbone. After the input is passed through this backbone, multi-scale features are obtained. We use the bottom feature to connect to the decoder. The decoder consists of the stacking of convolutional layers to recover the size of the feature maps. We use skip connect to merge the last three feature maps from the backbone to their corresponding feature maps. At the last two layers, we also add an attention module that contains both spatial and channel attention and an enhanced module \cite{qu2019enhanced,chen2020jstasr,yang2021LAFFNet} to refine the features. Finally, the decoder outputs the three-channel tensor referred to as the relit image. The model is illustrated in \figref{fig:architecture_SSS}. The details of this method are thoroughly explained in~\cite{yang2021S3Net}.

\subsection{iPAL-RelightNet: Any-to-any Multiscale
Intrinsic-Direct RelightNet (AMIDR-Net)}
\begin{figure}[ht]
    \centering
    \includegraphics[width=.48 \textwidth]{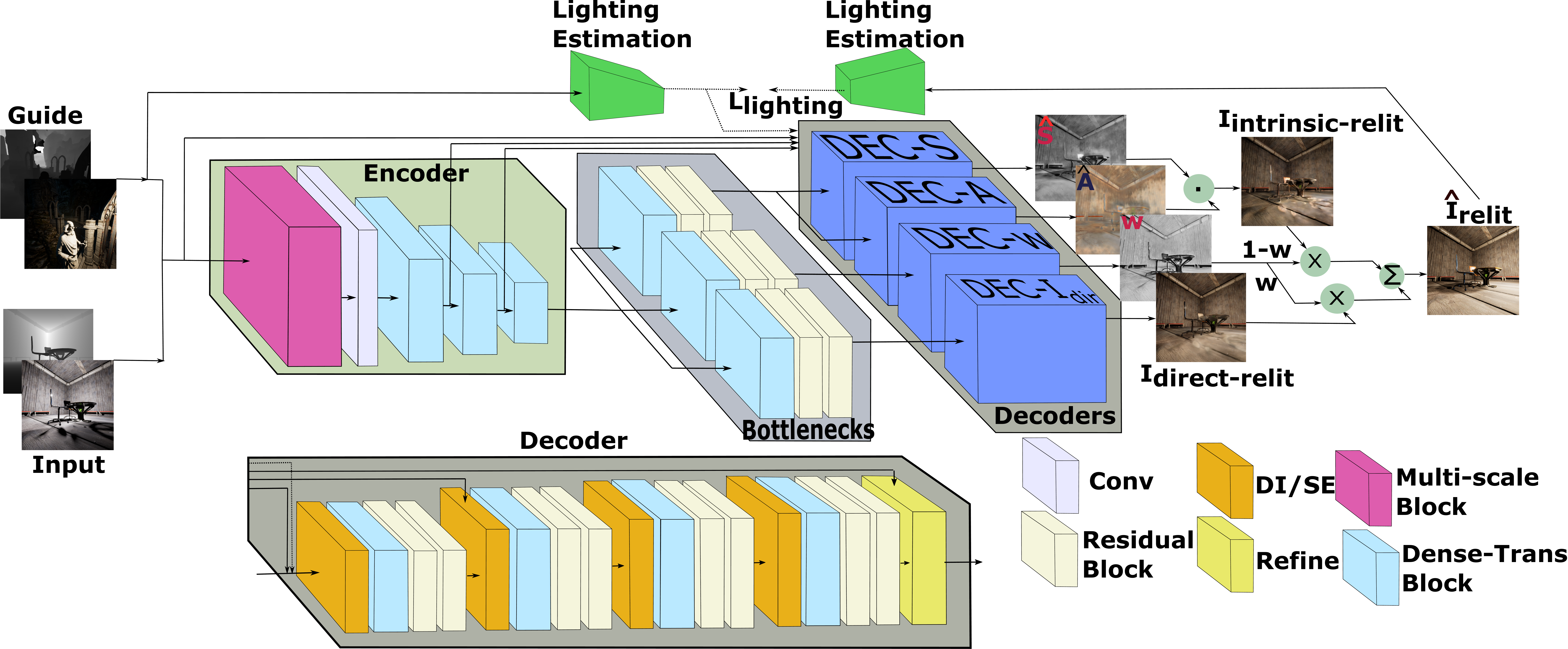}
    \caption{AMIDR-Net architecture overview.}
    \label{fig:AMIDR}
\end{figure}

Similar to one-to-one relighting, the proposed method \cite{Amir21} makes use of two strategies (intrinsic decomposition and black box) to generate the final relit output. The proposed network architecture (Fig. \ref{fig:AMIDR}) is based on the U-Net \cite{u-net}, while the encoder exploits the DenseNet \cite{densenet} pretrained layers. AMIDR-Net is different from OIDDR-Net in that it benefits from a multiscale block \cite{multiscale} for extracting more representative features from the input and the guide. Additionally, a lighting estimation network, which is trained separately over the training set and the corresponding illumination parameters, is incorporated first to provide the decoders with the illumination features of the guide, and second to compare the illumination of the relit output and the guide (via calculating a loss term: $L_{\textit{Lighting}}$). 

\subsection{IPCV IITM: Deep Residual Network for Image Relighting (DRNIR)}
\begin{figure}
    \centering
    \includegraphics[width = 0.45\textwidth]{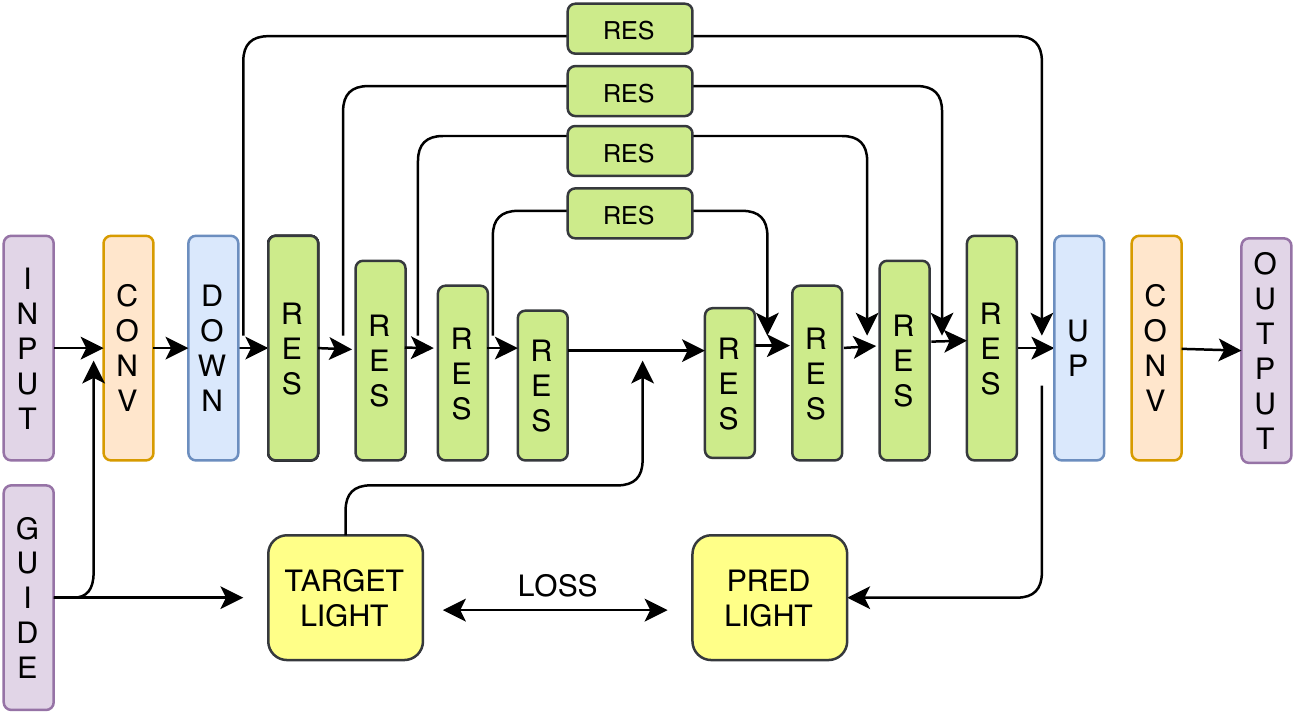}
    \caption{DRNIR network architecture overview.}
    \label{fig:ipcv_arch}
\end{figure}

We illustrate in Fig.~\ref{fig:ipcv_arch} the structure of the proposed residual network with skip connections, based on the  hourglass network \cite{zhou2019deep}. The network has an encoder-decoder structure \cite{purohit2019depth,suin2020degradation,suin2020spatially} with skip connections similar to \cite{huang2017densely}. Residual blocks are used in the skip connections, and Batch-Norm and ReLU non-linearity are used in each of the blocks. The encoder features are concatenated with the decoder features of the same level. The network takes the input image and directly produces the target image. Our solution converts the input RGB images to LAB for better processing. To reduce
the memory consumption without harming the performance, the team uses a pixel-shuffle block \cite{shi2016real} to downsample the image. In addition, the depth map is concatenated with the input image. The network is fist trained using the $\ell 1$ loss, then fine-tuned with the MSE loss. Experiments with adversarial loss did not lead to stable training. The learning rate of the Adam
optimizer is 0.0001 with a decay cycle of 200 epochs. Data augmentation is used to make the network more robust.

\subsection{Wit-AI-lab : Self-calibrated Relighting Network for any-to-any Relighting (SRN)}
\begin{figure}[t]
	\centering{\includegraphics[width=\linewidth]{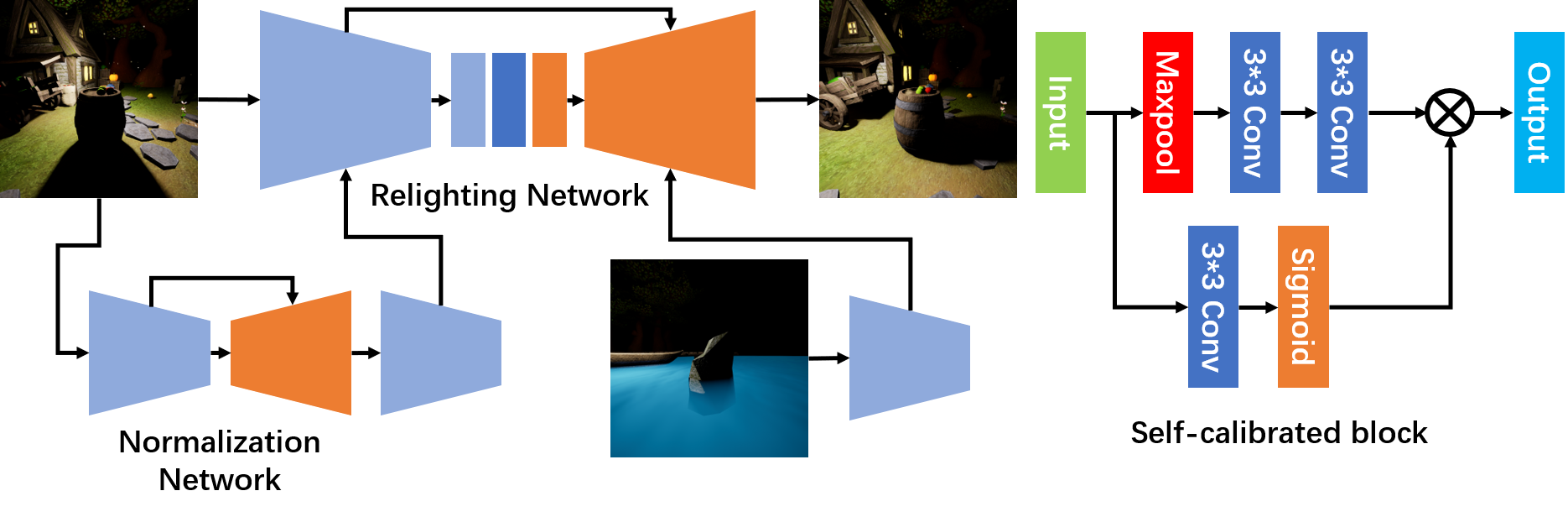}}
	\caption{Overall architecture of the SRN.}
	\label{fig:SRN_Network}
\end{figure}

We show in Fig.~\ref{fig:SRN_Network} the architecture of the proposed SRN method~\cite{wang2021multiscale} contains a normalization network and a relighting network. The normalization network aims to produce uniformly-lit white-balanced images, and the relighting network uses the latent feature representation of the guide image and uniformly-lit image produced by the normalization network to re-render the target image.
Considering that any-to-any relighting is a task of recalibrating the light source settings, the team proposes a novel self-calibrated block as the basic block of the feature encoder to calibrate the feature information for the normalization network and the relighting network iteratively, thereby improving the performance of any-to-any relighting.

\subsection{NPU-CVPG : Shadow Guided Refinement Network for Any-to-any Relighting}
Our proposed framework uses the self attention auto-encoder (SA-AE) of Hu \textit{et al.} \cite{10.1007/978-3-030-67070-2_32} as its backbone architecture. Two additional modules are added; namely a shadow synthesis module, and a shadow refinement module. The shadow synthesis module is designed as a conditional auto-encoder where the illumination direction is used as the conditional variable to control the depth for the shadow image generation process. This is done by first transforming the input image's depth and the predicted illumination direction into a common feature space, then fusing those features using an element-wise sum. The shadow refinement module is a computational unit used to further refine the relighting features. The entire framework is trained end-to-end. 

\section{Conclusion}
We presented the competition methods along with the benchmark results for our one-to-one and any-to-any depth guided image relighting challenges. The participating teams obtained high-quality results that are very competitive across the top-scoring solutions. We noted a significant improvement in the overall results compared to our first edition, and this was the case for all participating teams. This significant improvement strongly supports the use of depth information in relighting tasks. Hence, we strongly encourage future work to exploit depth information, and generally scene geometry information, for image relighting.

\appendix

\section*{Acknowledgements}
\addtocounter{footnote}{-2}
We thank the NTIRE 2021 sponsors: Huawei,
Facebook Reality Labs, Wright Brothers Institute, MediaTek, OPPO and ETH Zurich (Computer Vision Lab). %\url{https://data.vision.ee.ethz.ch/cvl/ntire21/}. 
We also note that all tracks were supported by the CodaLab infrastructure. 
% \url{https://competitions.codalab.org}.

\section{Teams and affiliations} \label{sec:teams}
% The teams who competed in the final test phase and have confirmed submissions are listed below in alphabetical order.\\

\noindent \textbf{NTIRE challenge organizers}\\
{\textit{Members}}: Majed  El  Helou,  Ruofan  Zhou,  Sabine  S\"usstrunk (\textit{\{majed.elhelou,sabine.susstrunk\}@epfl.ch}, EPFL, Switzerland), and Radu Timofte (\textit{radu.timofte@vision.ee.ethz.ch}, ETH Z\"urich, Switzerland).\\

\noindent \textbf{{-- AICSNTU-MBNet --}}\\
{\textit{Members}}: Hao-Hsiang Yang (\textit{islike8399@gmail.com}), Wei-Ting Chen, Hao-Lun Luo, Sy-Yen Kuo. \\
{\textit{Affiliation}}: ASUS Intelligent Cloud Services, ASUSTeK Computer Inc; Graduate Institute of Electronics Engineering, National Taiwan University; Department of Electrical Engineering, National Taiwan University. \\

\noindent \textbf{{-- AICSNTU-SSS --}}\\
{\textit{Members}}: Hao-Hsiang Yang (\textit{islike8399@gmail.com}), Wei-Ting Chen, Sy-Yen Kuo. \\
{\textit{Affiliation}}: ASUS Intelligent Cloud Services, ASUSTeK Computer Inc; Graduate Institute of Electronics Engineering, National Taiwan University; Department of Electrical Engineering, National Taiwan University. \\

\noindent \textbf{{-- alphaRelighting --}}\\
{\textit{Members}}: Chenghua Li (\textit{lichenghua2014@ia.ac.cn}), Bosong Ding, Wanli Qian, Fangya Li.\\
{\textit{Affiliation}}: Institute of Automation, Chinese Academy of Sciences, China. \\

\noindent \textbf{{--Couger AI--}}\\
{\textit{Members}}:Sabari Nathan (\textit{sabari@couger.co.jp}), Priya Kansal.\\
{\textit{Affiliation}}: Couger Inc.\\

\noindent \textbf{{-- DeepBlueAI --}}\\
{\textit{Members}}: Zhipeng Luo (\textit{luozp@deepblueai.com}), Zhiguang Zhang, Jianye He. \\
{\textit{Affiliation}}: DeepBlue Technology (Shanghai) Co., Ltd, China. \\

\noindent \textbf{{-- iPAL-RelightNet --}}\\
{\textit{Members}}: Amirsaeed Yazdani (\textit{amiryazdani@psu.edu}), Tiantong Guo, Vishal Monga.\\
{\textit{Affiliation}}:\\ 
The Pennsylvania State University\\
School of Electrical Engineering and Computer Science\\
The Information Processing and Algorithms Laboratory\\
(iPAL).\\

\noindent \textbf{{-- IPCV\_IITM --}}\\
{\textit{Members}}: Maitreya Suin (\textit{maitreyasuin21@gmail.com}), A. N. Rajagopalan. \\
{\textit{Affiliation}}:  Indian Institute of Technology Madras, India. \\

\noindent \textbf{{-- MCG-NKU --}}\\
{\textit{Members}}: Zuo-liang Zhu (\textit{nkuzhuzl@gmail.com}), Zhen Li, Jia-Xiong Qiu, Zeng-Sheng Kuang,  Cheng-Ze Lu, Ming-Ming Cheng, Xiu-Li Shao.\\
{\textit{Affiliation}}:  Nankai University, China. \\

\noindent \textbf{{-- NPU-CVPG --}}\\
{\textit{Members}}: Ntumba Elie Nsampi (\textit{elientumba@mail.nwpu.edu.cn}), Zhongyun Hu, Qing Wang.\\
{\textit{Affiliation}}:\\ 
Computer Vision and Computational Photography Group, School of Computer science, Northwestern Polytechnical University, China.\\

\noindent \textbf{{-- NTUAICS-ADNet --}}\\
{\textit{Members}}:  Wei-Ting Chen (\textit{f05943089@ntu.edu.tw}), Hao-Hsiang Yang, Hao-Lun Luo, Sy-Yen Kuo. \\
{\textit{Affiliation}}: Graduate Institute of Electronics Engineering, National Taiwan University; ASUS Intelligent Cloud Services, ASUSTeK Computer Inc; Department of Electrical Engineering, National Taiwan University. \\

\noindent \textbf{{-- NTUAICS-VGG --}}\\
{\textit{Members}}:  Hao-Lun Luo (\textit{r08921051@ntu.edu.tw}), Hao-Hsiang Yang, Wei-Ting Chen, Sy-Yen Kuo. \\
{\textit{Affiliation}}: Department of Electrical Engineering, National Taiwan University; ASUS Intelligent Cloud Services, ASUSTeK Computer Inc; Graduate Institute of Electronics Engineering, National Taiwan University. \\

\noindent \textbf{{-- usuitakumi --}}\\
{\textit{Members}}: Tongtong Zhao (\textit{daitoutiere@gmail.com}), Shanshan Zhao. \\
{\textit{Affiliation}}: Dalian Maritime University; China Everbright Bank. \\

\noindent \textbf{{-- VUE --}}\\
{\textit{Members}}: Fu Li (\textit{lifu@baidu.com}), Ruifeng Deng, Tianwei Lin, Songhua Liu, Xin Li, Dongliang He. \\
{\textit{Affiliation}}: Department of Computer Vision (VIS), Baidu Inc. \\

\noindent \textbf{{-- Wit-AI-lab --}}\\
{\textit{Members}}: Yuanzhi Wang (\textit{w906522992@gmail.com}), Tao Lu, Yanduo Zhang, Yuntao Wu. \\
{\textit{Affiliation}}: Hubei Key Laboratory of Intelligent Robot, Wuhan Institute of Technology, China. \\

{\small
\bibliographystyle{IEEEtran}
\bibliography{egbib}
}

\end{document}